\pdfoutput=1
\documentclass[twocolumn]{article}

\usepackage[utf8]{inputenc}
\usepackage[margin=1.25in]{geometry}
\usepackage{authblk}
\usepackage{setspace}
\usepackage{graphicx}
\graphicspath{{./}}
\usepackage{subcaption}
\usepackage{booktabs}
\usepackage{xurl}
\usepackage{hyperref}
\hypersetup{hidelinks}
\usepackage{amsmath}
\usepackage{amssymb}
\usepackage{float}
\usepackage{xcolor}
\usepackage{colortbl} 
\usepackage{microtype}
\usepackage{graphicx}
\usepackage{subcaption}
\usepackage{multirow} 
\usepackage{xcolor} 
\usepackage{amsmath}
\usepackage{amssymb}
\usepackage{mathtools}
\usepackage{amsthm}
\usepackage{tabularx}
\usepackage{placeins}
\usepackage{bm}
\usepackage{chngcntr}
\usepackage{pgfplots}
\usepgfplotslibrary{groupplots}
\usetikzlibrary{positioning,calc}
\pgfplotsset{compat=1.17}
\definecolor{AnormColor}{RGB}{31,119,180}
\definecolor{AstarColor}{RGB}{214,39,40}

\newcommand{\first}[1]{\cellcolor{red!40}\textbf{#1}}
\newcommand{\second}[1]{\cellcolor{orange!40}\textbf{#1}}
\newcommand{\third}[1]{\cellcolor{green!40}\textbf{#1}}

\usepackage[capitalize,noabbrev]{cleveref}

\usepackage[switch]{lineno}       
\usepackage[numbers,sort&compress]{natbib}

\title{Geometric Evolution Graph Convolutional Networks:\\
  Enhancing Graph Representation Learning via Ricci Flow}

\author[1]{Jicheng Ma}
\author[1]{Yunyan Yang}
\author[1]{Juan Zhao}
\author[2*]{Liang Zhao}

\affil[1]{School of Mathematics, Renmin University of China, Beijing, China.}
\affil[2]{School of Mathematical Sciences, Key Laboratory of Mathematics and Complex Systems of MOE, Beijing Normal University, Beijing, China.}
\affil[*]{*Corresponding author: liangzhao@bnu.edu.cn}

\date{}

\begin{document}
\twocolumn[
\begin{@twocolumnfalse}

\maketitle

\begin{abstract}
  Graph geometry provides structural information for message passing, but using a single curvature snapshot or only the output of curvature-guided rewiring leaves the ordered evolution of geometry largely unexploited. We introduce the Geometric Evolution Graph Convolutional Network (GEGCN), a framework for graph representation learning that explicitly models geometric evolution on weighted graphs. First, a geometric evolution generator updates edge weights via discrete Ricci flow to produce an ordered sequence of curvature--weight pairs. A Long Short-Term Memory (LSTM) encoder then maps each edge sequence to an adaptive importance score. Finally, the normalized scores define a geometry-aware graph convolution operator on the original adjacency support. This design incorporates evolving geometric context into message passing while preserving graph connectivity. Extensive experiments demonstrate that GEGCN achieves excellent performance on node classification tasks. We test across 17 benchmark datasets, covering homophilic, traditional and filtered heterophilic, and large-scale settings. GEGCN attains an average rank of 1.76 among compared methods and reaches a test accuracy of 75.76\% on OGBN-arXiv. Ablations show the benefit of the ordered trajectory over the evaluated single-state, pooling, non-recurrent, and order-perturbed alternatives. These results support ordered geometric evolution encoding as an effective approach to graph representation learning.
\end{abstract}


\vspace{1em}
\end{@twocolumnfalse}
]


\section{Introduction}

Graph Neural Networks (GNNs), especially Graph Convolutional Networks (GCNs) and their variants, have achieved remarkable success for learning representations on graph-structured data by iteratively propagating and aggregating information from local neighborhoods~\citep{zhou2020gnnsurvey}. Despite their effectiveness, most existing GNNs, such as GCN~\citep{kipf2017semi}, GAT~\citep{velivckovic2018graph}, and GraphSAGE~\citep{hamilton2017inductive}, treat graphs as static snapshots and focus on local message passing. This paradigm often overlooks the rich, global geometric properties inherent in graph structures. 

Recently, the concept of discrete curvature~\citep{forman2003bochner,ollivier2009ricci,lin2011ricci} has been successfully adapted to network analysis. An expanding corpus of research confirms its utility as a principled way to characterize local and global structural features of graphs, such as bottlenecks~\citep{topping2022oversquash}, community structure~\citep{ni2019community_ricciflow,lai2022normalized_ricciflow}, and network fragility~\citep{sandhu2016market_ricci}. This geometric perspective has naturally inspired notable efforts to integrate discrete curvature into GNN pipelines \citep{ye2020curvature,topping2022oversquash,nguyen2023revisiting,chen2025gnrf,wang2026adaptive}. These approaches demonstrate the usefulness of geometry, but they generally rely on static, precomputed curvature, or a topology produced by a curvature-based rewiring rather than learning directly from the complete ordered trajectory of a discrete Ricci flow.

Discrete Ricci flow~\citep{ollivier2009ricci,bauer2012ollivier,bai2024ollivier,ma2024modified} deforms the weighted graph metric by recomputing curvature and updating edge weights in each iteration. It therefore produces an ordered sequence of geometric states rather than a static snapshot. This evolution trajectory contains discriminative structural information, and explicitly encoding this sequence enables learning more expressive node representations. This motivates the scientific question:

How can the ordered geometric trajectory generated by discrete Ricci flow be incorporated into graph representation learning, rather than using only a single snapshot of graph geometry or only the graph produced by a curvature-based rewiring?

In parallel, recurrent neural networks (RNNs), especially Long Short-Term Memory (LSTM) networks~\citep{hochreiter1997lstm}, have demonstrated strong capability in modeling temporal dependencies and structured sequences. This property makes LSTMs a suitable candidate for processing and aggregating the ordered geometric evolution trajectory considered here.

Motivated by these observations, we propose the Geometric Evolution Graph Convolutional Network (GEGCN), a novel framework that integrates Ricci flow with deep graph models by explicitly modeling geometric evolution as a sequential process. Specifically, we generate a sequence of evolving graphs via discrete Ricci flow, encode each edgewise curvature--weight sequence with an LSTM, map the final encoding to an edge importance score, and use the normalized score matrix for graph convolution. In this way, GEGCN bridges discrete differential geometry and graph neural networks through a dynamic representation learning paradigm.

We evaluate GEGCN across 17 datasets covering homophilic, unfiltered/filtered heterophilic, and large-scale settings. GEGCN obtains an average rank of 1.76 across all datasets. Targeted ablations show that the full ordered trajectory is more effective than the evaluated single-state, pooling, non-recurrent, shuffled, and reversed alternatives. These results support GEGCN's competitive performance and suggest a promising direction for unifying geometric analysis and graph neural networks.

Our contributions are summarized as follows:

\begin{itemize}
	\item {\bf A New Geometric Perspective}. We introduce discrete Ricci flow into deep graph representation learning, providing a principled method to model the geometric evolution of graph structures.
	\item {\bf A Simple Architecture}. The proposed GEGCN maintains a concise and elegant design by integrating discrete Ricci flow, an LSTM encoder, and graph convolution in a three-stage architecture. The encoder maps Ricci-flow geometric states to propagation weights, allowing the model to incorporate flow-time geometric information without modifying the original edge support.
	\item {\bf Broad Empirical Validation}. Extensive experiments across datasets of varying sizes, homophilic/heterophilic settings, and filtered benchmarks demonstrate the competitive performance of flow-based graph representation learning.
\end{itemize}

\section{Related Works}

Our work lies at the intersection of geometry of graphs, neural differential equations, and graph neural networks. We categorize related research into two primary strands, highlighting their distinctions from our proposed GEGCN.

\subsection{Geometric Graph Neural Networks}

A significant line of research enhances GNNs by incorporating discrete curvature or related metrics to modulate message passing. For instance, CurvGN~\citep{ye2020curvature} and CGNN~\citep{li2022cgcn} explicitly use Ollivier-Ricci curvature to reweight message passing, demonstrating improved performance. $\kappa$HGCN~\citep{yang2023hgcn} utilizes a curvature-aware hyperbolic graph convolutional neural network in order to guide message passing on tree-like structures. SelfRGNN~\citep{sun2022selfrgnn} tackles temporal graphs by formulating a time-dependent functional curvature to reweight message passing. RC-UFG~\citep{shi2023curvature} introduces two curvature-based framelet models, namely RC-UFG (Hom) and RC-UFG (Het), showcasing their enhanced adaptability on homophilic and heterophilic graph datasets.

It is worth noting that \cite{topping2022oversquash} formally established the connection between over-squashing phenomena in GNNs and graph curvature. SDRF proposed in \cite{topping2022oversquash}, is a curvature-based rewiring method that modifies graph topology, which the authors themselves note is ``different from more direct extensions of Ricci flow on graphs''. Works \cite{nguyen2023revisiting,giraldo2023tradeoff,fesser2024mitigating} follow a similar intent to \cite{topping2022oversquash} in their use of curvature. In this direction, CurvDrop~\citep{liu2023curvdrop} introduces a curvature-based dropout sampling technique to mitigate over-smoothing and over-squashing. CurvGIB~\citep{fu2025discrete} advances this by applying a variational information bottleneck principle, aiming to learn optimal information transport patterns for better node representations. ARGNN instead learns a node-adaptive anisotropic Riemannian metric with a Ricci-flow-inspired regularizer~\citep{wang2026adaptive}. These methods provide relevant comparisons, but they do not encode the ordered geometric trajectory as GEGCN.

\subsection{Neural Diffusion Equations on Graphs}

Another category frames deep learning on graphs as diffusion processes. GRAND~\citep{chamberlain2021grand} and Beltrami flow~\citep{chamberlain2021beltrami} model feature propagation with diffusion equations, providing a principled, continuous-depth perspective on GNNs. Specifically, Beltrami flow employs the Laplace-Beltrami operator within a diffusion partial differential equation (PDE), making it a non-Euclidean diffusion process where node features are augmented with topology-derived positional encodings.

Closely related is GNRF~\citep{chen2025gnrf}, which conceptualizes node feature learning as a curvature-driven diffusion. GNRF essentially implements a learned, curvature-informed diffusion PDE rather than an explicit discrete Ricci flow that evolves the graph structure itself to generate a geometric sequence.

\subsection{Summary and Positioning}

Using curvature to guide message passing in GNNs has emerged as a well-recognized and natural idea in recent years. While existing geometric GNNs have validated the utility of curvature for enhancing model performance, GEGCN is a concise, principled design that differs from existing curvature-based approaches.

Existing works that incorporate curvature into GNNs typically rely on a static snapshot of curvature, whether precomputed or learned as a fixed parameter. This captures only a single-scale geometric view and fails to model the multi-scale geometric states of graphs. Curvature-guided rewiring methods such as SDRF may also adopt iterative Ricci flow updates, but their output is a modified topology rather than a learned encoding of every intermediate state.

We specify that \emph{multi-scale geometric state} (see Sec. \ref{sec:flow}) specifically refers to the diverse geometric states obtained throughout Ricci flow evolution.

While PDE-based diffusion models are powerful, they typically do not incorporate the geometric evolution enabled by discretized Ricci flow. In GNRF, curvature is predicted by an auxiliary neural module (EdgeNet). As the authors of GNRF note, EdgeNet can be viewed as approximating Ollivier-Ricci, Forman-Ricci, or other curvature forms. GNRF employs EdgeNet to predict coefficients in a feature-diffusion model, rather than encoding an explicit geometric metric.

In contrast, our GEGCN evolves graph geometry via discrete Ollivier--Ricci flow and integrates the resulting dynamic geometric information into a standard GCN using an LSTM encoder, thereby capturing the full trajectory of geometric change. This pipeline retains the original graph adjacency structure and builds upon the geometric analysis framework originally extended to metric spaces by \cite{ollivier2009ricci}.

\section{Preliminaries}

This section introduces the foundational concepts for our work. We first establish key notations and review the basics of graph neural networks. We then present the notions of discrete Ricci curvature and Ricci flow on graphs. Finally, we discuss RNNs and the LSTM architecture.

\subsection{Notations and Graph Convolutional Networks}

Let $\mathcal{G} = (\mathcal{V},\mathcal{E})$ denote an undirected graph, where $\mathcal{V} = \{v_1, \dots, v_N\}$ is the set of $N$ nodes and $\mathcal{E} \subseteq \mathcal{V} \times \mathcal{V}$ is the set of edges. The graph structure is represented by an adjacency matrix $\mathbf{A} \in \{0,1\}^{N \times N}$, where $\mathbf{A}_{ij} = 1$ if $(v_i,v_j)\in\mathcal{E}$ and $\mathbf{A}_{ij}=0$ otherwise. Each node $v_i$ is associated with a feature vector $\mathbf{x}_i \in \mathbb{R}^d$, and the node feature matrix is denoted as $\mathbf{X} \in \mathbb{R}^{N \times d}$. The neighborhood of node $v_i$ is defined as $\mathcal{N}(i) = \{v_j \in \mathcal{V} : (v_i, v_j) \in \mathcal{E}\}$. The degree of $v_i$ is $d_i = |\mathcal{N}(i)|$, and the degree matrix is $\mathbf{D} = \mathrm{diag}(d_1, \dots, d_N)$.

Graph neural networks learn node representations by iteratively aggregating information from local neighborhoods, a paradigm known as message passing. As a foundational and widely adopted instance, the graph convolutional network layer implements a specific form of this mechanism. The layer-wise propagation rule of a GCN is defined as
\begin{equation}
	\label{gcnlayer}
	\mathbf{H}^{(\ell+1)} = \sigma\!\left(\hat{\mathbf{A}} \mathbf{H}^{(\ell)} \mathbf{W}^{(\ell)}\right),
\end{equation}
where
$\mathbf{H}^{(\ell)} \in \mathbb{R}^{N \times h_\ell}$ is the matrix of hidden node representations at layer $\ell$, with $\mathbf{H}^{(0)} = \mathbf{X}$, $\mathbf{W}^{(\ell)} \in \mathbb{R}^{h_\ell \times h_{\ell+1}}$ is a layer-specific learnable weight matrix, $\sigma(\cdot)$ denotes a non-linear activation function, and $\hat{\mathbf{A}}$ is the symmetrically normalized adjacency matrix
\begin{equation}
	\label{adjacencymatrix}
	\hat{\mathbf{A}} = \tilde{\mathbf{D}}^{-\frac{1}{2}} \tilde{\mathbf{A}} \tilde{\mathbf{D}}^{-\frac{1}{2}}.
\end{equation}
Here, $\tilde{\mathbf{A}}=\mathbf{A}+\mathbf{I}_N$, $\mathbf{I}_N$ is the $N \times N$ identity matrix, and $\tilde{\mathbf{D}} = \text{diag}\left(\sum_j\tilde{\mathbf{A}}_{ij}\right)$. This normalization stabilizes the scale of feature propagation across nodes with varying degrees and can be interpreted as applying a localized spectral filter on the graph.

After stacking $L$ such layers, the final output $\mathbf{H}^{(L)}$ provides the node representations used for downstream tasks, such as node classification.

\subsection{Discrete Ricci Curvature and Ricci Flow}
\label{sec:flow}

Discrete curvature discretizes notions from differential geometry to quantify the deviation of a graph's local structure from flatness. Among several discrete analogues of curvature~\citep{Menger1930,haantjes1947discrete,bakry1985hyper,lin2011ricci}, the Ollivier-Ricci curvature~\citep{ollivier2009ricci} is widely used, providing a probabilistic interpretation via optimal transport. For an edge $e = (u, v)$, its curvature $\kappa_e$ is defined as
$\kappa_e = 1 - \frac{W_1(m_u, m_v)}{\rho_{e}}$, where $\rho_{e}$ is the shortest-path distance between nodes $u$ and $v$, and $W_1(m_u, m_v)$ is the $1$-Wasserstein distance between two probability measures $m_u$ and $m_v$ defined on the neighborhoods of $u$ and $v$, respectively. Intuitively, positive curvature indicates locally clustered structures, zero curvature suggests flat regions, and negative curvature corresponds to tree-like or bottleneck structures.

On a graph $\mathcal{G}$ with initial edge weights $\{w_e\}_{e \in \mathcal{E}}$, a discrete Ricci flow can be defined to evolve these weights based on edge curvatures. For an edge $e = (u, v)$ with curvature $\kappa_e(t)$ and weight $w_e(t)$ at iteration $t$, the continuous flow is described by
$w'_e(t) = -\kappa_e(t) \rho_{e}(t)$. In practice, we adopt a discretized version with a step size $\delta > 0$
\begin{equation}
	\label{discreteRF} w_e(t+1) = w_e(t) - \delta \kappa_e(t) \rho_{e}(t).
\end{equation}

Repeating this process $T$ times, the flow generates a sequence of weighted graphs $\{\mathcal{G}(t)\}_{t=0}^{T}$. Note that, from the definitions of $\kappa_e$ and $\rho_e$, whenever $\delta\leq 1$, the iteratively updated weights $w_e(t+1)$ remain strictly positive, provided that the initial weights $w_e(0)$ are positive.

Let $\mathbf w(t)=(w_e(t))_{e\in\mathcal E}$, and let $\bm\kappa(t)=(\kappa_e(t))_{e\in\mathcal E}$. We call $\mathcal S_t=(\mathbf w(t),\bm\kappa(t))$
the Ricci-flow geometric state at flow time $t$. The states share the same adjacency support but differ in edge weights, curvature, and the shortest-path metric induced by $\mathbf w(t)$.

Accordingly, \emph{multi-scale} in this work refers to encoding $\mathcal S_t$ from multiple Ricci-flow times, but the term does not imply direct node-feature aggregation over multiple hop ranges. For an edge $e=(u,v)$, the random-walk measures used by Ollivier--Ricci curvature depend on the current weights incident to $u$ and $v$. Repeated simultaneous updates can therefore propagate dependence through adjacent edges. In addition, the costs used by optimal transport may depend on remote edges lying on the active shortest path. This expands the range of possible structural dependence of $\mathcal{S}_t$ along the flow. 

More details regarding discrete Ricci curvature and Ricci flow are provided in Appendix \ref{app:riccicurvature}. A dependency statement and controlled verification of Ricci-flow geometric states are given in Appendix~\ref{app:dependency}.

\subsection{Recurrent Neural Networks and Long Short-Term Memory}

RNNs offer a natural framework for modeling sequential data, enabling them to process the dynamic graph sequences. However, standard RNNs suffer from the vanishing gradient problem, where gradients diminish exponentially during backpropagation, severely limiting their ability to learn long-range dependencies within a sequence.

LSTMs introduce a gated cell state that enables selective retention and forgetting of information across long sequences. The forget gate determines what information from the previous cell state should be retained or discarded. The input gate regulates the incorporation of new candidate information. The output gate controls which aspects of the updated cell state are propagated to the hidden state. This gating mechanism enables LSTMs to maintain stable gradients over long sequences and selectively preserve long-range dependencies while flexibly integrating new inputs.

This capability makes LSTMs particularly suitable for modeling the geometric evolution sequences produced by discrete Ricci flow. Repeated flow updates cause the state trajectory to depend on geometric information from an expanding range, as stated in Sec. \ref{sec:flow}. The LSTM encodes these Ricci-flow geometric states, and the resulting encoded trajectory is utilized to construct propagation weights for the subsequent graph convolution. While the formal node-feature receptive field remains governed by graph convolution depth, conditioning local weights on potentially broader geometric context can enhance the effective utilization of this fixed receptive field. This design yields richer and more expressive node representations than those obtained from standard GCNs operating on static graph snapshots.

\begin{figure*}[htbp]
	\centering
	\includegraphics[trim={5mm 65mm 20mm 10mm}, clip, width=15cm]{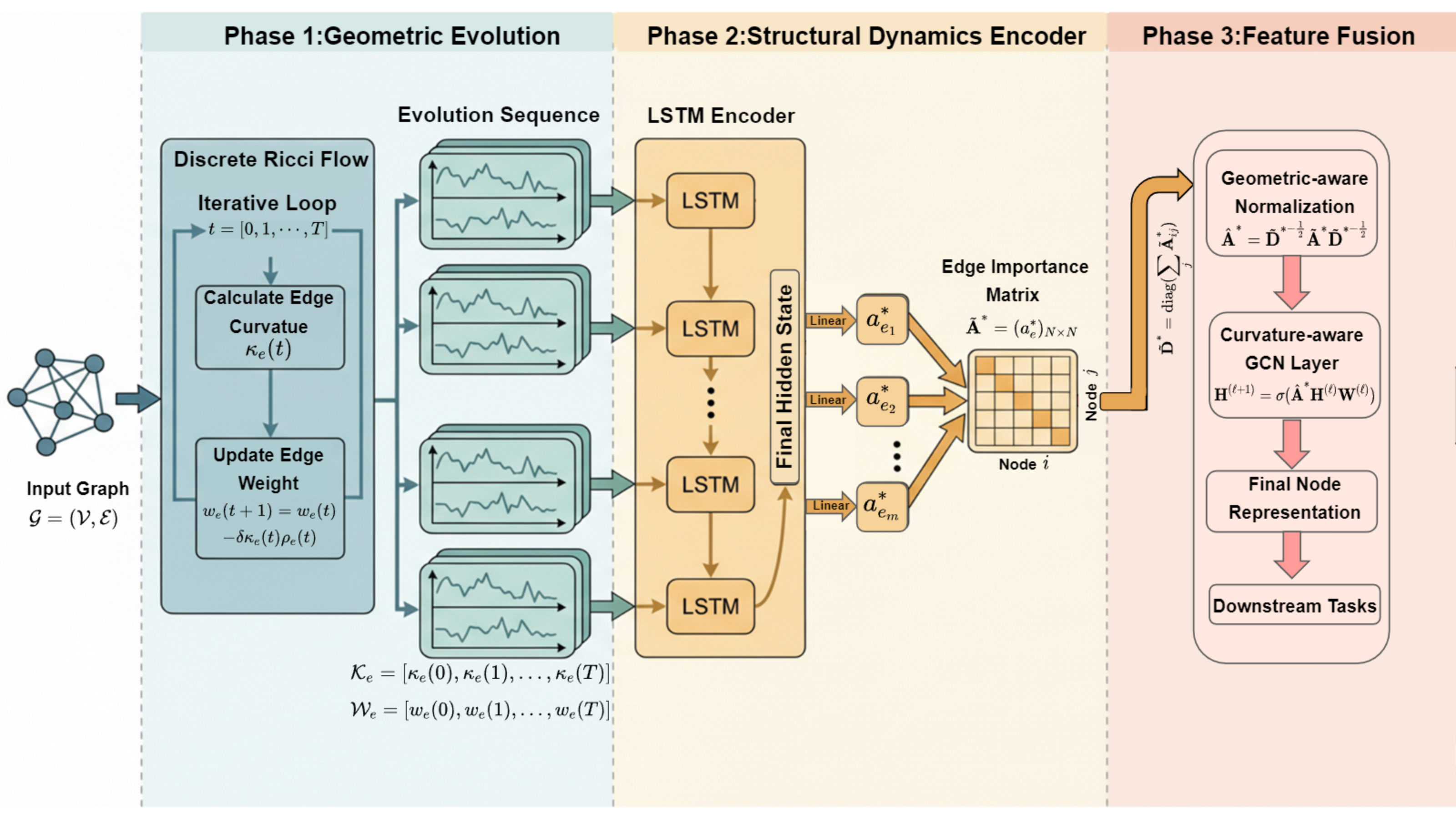}
	\caption{\textbf{Overview of the GEGCN framework.} It comprises three phases: (i) Geometric Evolution Generator via Discrete Ricci Flow; (ii) Structural Dynamics Encoder for encoding geometric states and constructing the edge importance matrix via LSTM; and (iii) Curvature-aware Graph Convolution that normalizes the edge importance matrix and propagates node features.}
	\label{fig:flow chart}
\end{figure*}

\section{Methodology}

The proposed Geometric Evolution Graph Convolutional Network is designed to encode the Ricci-flow geometric states by integrating discrete Ricci flow with sequential deep learning. GEGCN comprises three key components (Figure~\ref{fig:flow chart}): (i) a Geometric Evolution Generator that produces a graph sequence via discrete Ricci flow; (ii) a Structural Dynamics Encoder based on LSTM; and (iii) a Curvature-aware Graph Convolution module for learning final node representations. 

\subsection{Geometric Evolution Generator via Discrete Ricci Flow}
\label{generator}

This module generates the Ricci-flow geometric states of the input graph via discrete Ricci flow. Given an input graph $\mathcal{G}=(\mathcal{V},\mathcal{E})$, each edge $e \in \mathcal{E}$ is assigned an initial weight $w_e(0)$, which is typically set to $1$.

We construct a temporal graph sequence using discrete Ricci flow. The process begins by computing the initial discrete Ricci curvature $\kappa_e(0)$ for every edge $e \in \mathcal{E}$. We then perform $T$ iterations of the discrete Ricci flow. In each iteration $t$, after updating the edge weight $w_e(t+1)$ according to Eq.~\eqref{discreteRF}, we recalculate the corresponding Ricci curvature $\kappa_e(t+1)$ and the shortest-path distance $\rho_e(t+1)$ based on the updated weighted graph. This iterative procedure produces, for each edge $e$, two evolution sequences that encode its geometric dynamics. One is the curvature sequence
\[
\mathcal{K}_e= [\kappa_e(0), \kappa_e(1), \ldots, \kappa_e(T)],
\]
and another is the weight sequence
\[
\mathcal{W}_e= [w_e(0), w_e(1), \ldots, w_e(T)].
\]

Conceptually, while the classical Ricci flow on a smooth manifold defines a continuous geometric deformation that preserves fundamental structures, its discrete counterpart in Eq.~\eqref{discreteRF} induces a stepwise evolution of the graph, preserving its underlying adjacency topology.

\subsection{Structural Dynamics Encoder Based on LSTM}
\label{lstm}

To learn latent representations of the geometric dynamics, we design an LSTM-based encoder that processes the curvature and weight sequences generated in Sec.~\ref{generator}. After the Ricci-flow process on the original edge set $\mathcal{E}$ is completed, we append one self-loop to each node and assign to it a constant curvature sequence $[0,\cdots, 0]$ and a constant weight sequence $[1,\cdots, 1]$.

At time step $t$, the feature vector $\mathbf{x}_e^{(t)} = [\kappa_e(t), w_e(t)]$ of each original edge and appended self-loop is fed into the LSTM cell:
\begin{equation*}
	\mathbf{h}_e^{(t)}, \mathbf{c}_e^{(t)} = \text{LSTM}\left( \mathbf{x}_e^{(t)}, \mathbf{h}_e^{(t-1)}, \mathbf{c}_e^{(t-1)} \right),
\end{equation*}
where $\mathbf{h}_e^{(t)} \in \mathbb{R}^d$ denotes the hidden state, representing the accumulated geometric information up to the current time. The cell state $\mathbf c_e^{(t)}\in\mathbb R^d$ is the gated internal memory of the LSTM. It selectively retains information from earlier Ricci-flow steps while incorporating the current curvature and weight, allowing the encoder to summarize dependencies along the ordered sequence $\{[\kappa_e(t),w_e(t)]\}_{t=0}^{T}$. The full gate equations are given in Appendix~\ref{app:lstm}.

We input the final hidden state $\mathbf{h}_e^{(T)}$ of each edge into an edge-scoring head to output the importance score of the edge. This head consists of a linear transformation followed by a sigmoid activation $a^*_e = \sigma\left( \mathbf{W}_s \mathbf{h}_e^{(T)} + b_s \right)$, where $a^*_e \in (0, 1)$ is the learned edge importance score. By integrating $\mathcal{K}_e$ and $\mathcal{W}_e$ into the score $a^*_e$, this module can adaptively emphasize structural information that is persistent or critical across distinct geometric states.

\subsection{Curvature-aware Graph Convolution}

The classical GCN updates the node embedding matrix $\mathbf{H}^{(\ell)}$ to $\mathbf{H}^{(\ell+1)}$ according to Eq.~\eqref{gcnlayer}. The normalized adjacency matrix $\hat{\mathbf{A}}$ in GCNs relies on static node degrees via $\tilde{\mathbf{D}}$, treating all edges incident to nodes of the same degree in a uniform manner. This design fails to account for the nuanced geometric hierarchy within the graph.

Discrete Ricci flow provides a much richer, dynamic measure for each edge over time. To integrate this vital geometric information, we propose replacing the static, degree-based normalization matrix $\hat{\mathbf{A}}$ with a geometry-aware normalization matrix. Let $\tilde{\mathbf{A}}^*=(a^*_e)_{N\times N}$ be the edge importance matrix obtained by the LSTM encoder in Sec.~\ref{lstm}, and $\tilde{\mathbf{D}}^* = \text{diag}\left(\sum_j \tilde{\mathbf{A}}^*_{ij}\right)$. The geometry-aware normalization matrix is defined as
\[
\hat{\mathbf{A}}^*=\tilde{\mathbf{D}}^{*-\frac{1}{2}}\tilde{\mathbf{A}}^*\tilde{\mathbf{D}}^{*-\frac{1}{2}}.
\]
The update at the $\ell$-th layer of the curvature-aware graph convolution is defined as
\[
\mathbf{H}^{(\ell+1)} =  \sigma(\hat{\mathbf{A}}^* \mathbf{H}^{(\ell)} \mathbf{W}^{(\ell)}).
\]

We remark that the matrix $\tilde{\mathbf{A}}=\mathbf{A}+\mathbf{I}_N$ in Eq.~\eqref{adjacencymatrix} represents the adjacency matrix after adding a self-loop to every node in the graph. This ensures that each node retains its own features during the message-passing process of GCN. To align with this design, we append one self-loop to each node only after the Ricci-flow process in Sec.~\ref{generator} has been completed on the original edge set. Each appended self-loop is assigned a constant weight of $1$ and a constant curvature of $0$ at all time steps, and therefore does not affect the curvature or distance computation of the original edges. After normalization by $\tilde{\mathbf{D}}^*$, the self-loop score (the diagonal entry of $\hat{\mathbf{A}}^*$) is jointly determined by the scores of all incident edges of that node and may therefore vary across nodes. Consequently, self-feature retention adapts to the learned geometric context rather than being determined only by unweighted degree.

The connection among the three modules can be summarized as 
\[
\{[\kappa_e(t),w_e(t)]\}_{t=0}^{T}\rightarrow\mathbf h_e^{(T)}\rightarrow a_e^*\rightarrow\widehat{\mathbf A}^*.
\]
The geometric evolution generator produces the flow-time sequence, the LSTM encoder maps the sequence to an edge importance score, and the curvature-aware graph convolution uses the normalized score matrix to propagate node features.

\section{Experiments}
\subsection{Experimental Setup}\label{exp_data}
\paragraph{Datasets.}
We perform extensive experiments to validate the performance of GEGCN across a wide range of benchmark datasets, including homophilic/heterophilic graphs, filtered graphs, and large-scale graphs. 

For homophilic scenarios, we utilize three citation networks (Cora, Citeseer, and PubMed \citep{sen2008collective}), the Coauthor CS network based on the Microsoft Academic Graph from the KDD Cup 2016 challenge, and the Amazon Photos product network \citep{mcauley2015image}. Following the partitioning scheme in \cite{shi2023curvature}, we employ {\bf a uniform 20\%/10\%/70\% split for training, validation, and testing}.

To test robustness to violation of the homophily assumption, we extend our evaluation to datasets including Cornell, Texas, and Wisconsin from WebKB \citep{craven1998learning}, as well as Chameleon \citep{rozemberczki2021multi} and Actor \citep{tang2009social}. Note that, for these heterophilic benchmarks, we adopt the experimental protocol from \cite{topping2022oversquash}, {\bf using a 60\%/20\%/20\% data split}. {\bf The test set is always fixed rather than randomly sampled}, and the remaining nodes are then randomly divided into training and validation sets. This is to ensure that the test set is not used for any training or hyperparameter tuning prior to final evaluation.

For heterophilic scenarios, to further evaluate the model under challenging heterophilic settings and address the train-test data leakage issue reported in \cite{platonov2023criticallook}, we conduct additional experiments on six heterophilic benchmark datasets provided in \cite{platonov2023criticallook}, namely filtered Squirrel, filtered Chameleon, Roman-empire, Amazon-ratings, Minesweeper and Questions. For these datasets, we follow {\bf the same data split protocol as \cite{platonov2023criticallook}}.

Finally, to evaluate the large-scale performance and computational efficiency of GEGCN, we conduct experiments on the large-scale OGBN-arXiv benchmark and report both training and inference time. On this dataset, we adopt {\bf the standard train/validation/test split provided by OGB \citep{hu2020open}}. Specifically, we use papers published until 2017 as the training set, those in 2018 as the validation set, and those published in 2019 and later as the test set.

\paragraph{\textbf{Comparison Methods.}}
In the evaluation on homophilic datasets, we compare our model against a diverse set of competitive baselines ranging from classical GNNs to spectral-based methods. These include MLP, GCN \citep{kipf2017semi}, GAT \citep{velivckovic2018graph}, and GraphSAGE \citep{hamilton2017inductive}. We also include models designed for deeper or more flexible propagation, such as JKNet \citep{xu2018representation}, APPNP \citep{gasteiger2018predict}, GPR-GNN \citep{chien2020adaptive}, MoNet \citep{monti2017geometric}, and UFGConv \citep{zheng2021framelets}. Moreover, we consider the geometry-based approaches CurvGN \citep{ye2020curvature}, RC-UFG \citep{shi2023curvature}, and SDRF~\citep{topping2022oversquash}, along with BORF~\citep{nguyen2023revisiting}, which performs batch Ollivier-Ricci flow rewiring, AFRC~\citep{fesser2024mitigating}, which uses augmented Forman-Ricci curvature for scalable rewiring, and ARGNN~\citep{wang2026adaptive}, which learns node-adaptive Riemannian metrics.

For the traditional heterophilic benchmarks, we compare GEGCN against standard GNNs and several specialized architectures with distinct geometric or propagation mechanisms: (i) GCN as a fundamental backbone; (ii) DIGL \citep{gasteiger2019diffusion}, which employs diffusion to improve graph connectivity; (iii) +FA \citep{alon2020bottleneck}, which introduces a fully adjacent layer to bypass structural bottlenecks; (iv) geometry-based methods including SDRF\footnote{Note that reproducibility issues regarding the original results have been reported due to incomplete experimental details (\href{https://github.com/jctops/understanding-oversquashing/issues/3}{Issue}). To ensure a fair comparison with GEGCN, we reproduce the SDRF experiments using the protocol described in this study.} \citep{topping2022oversquash}, BORF~\citep{nguyen2023revisiting}, AFRC~\citep{fesser2024mitigating}, GNRF~\citep{chen2025gnrf}, and ARGNN~\citep{wang2026adaptive}.

For the six heterophilic datasets introduced in \cite{platonov2023criticallook}, we compare against multiple strong baselines employed in \cite{platonov2023criticallook} and GNRF \citep{chen2025gnrf}. Among them, ResNet \citep{he2016deep} and GAT \citep{velivckovic2018graph} are representative neural architectures. H2GCN \citep{zhu2020beyond}, GPR-GNN \citep{chien2020adaptive}, GloGNN \citep{li2022finding}, FSGNN \citep{maurya2022simplifying}, FAGCN \citep{bo2021beyond}, and GBK-GNN \citep{du2022GBK} are models designed for node classification under heterophily.

On the large-scale OGBN-arXiv dataset, we compare GEGCN with models including MLP, GCN \citep{kipf2017semi}, GAT \citep{velivckovic2018graph}, GraphSAGE \citep{hamilton2017inductive}, a random walk-based method Node2vec \citep{grover2016node2vec}, the strong baselines UniMP \citep{shi2021masked} and AGDN \citep{sun2025adaptive}, and the recent text-attributed graph model BiGTex~\citep{beiranvand2026integrating}.

For all the above benchmark datasets, the average results of 10 random experiments are reported, and the top {\color{red!60}{first}}, {\color{orange!70}{second}}, and {\color{green!40}{third}} are highlighted. Hyperparameters were selected by automated search over the predefined spaces in Appendix \ref{app:hyper_setting}, using the mean validation accuracy as the model-selection criterion.

\begin{table*}[htbp]
	\centering
	\caption{Mean test accuracy (\%) $\pm$ standard deviation across five homophilic benchmarks.}
	\label{tab:result_realworld_dataasets_5} 
	\small
	\begin{tabular}{lccccc} 
		\toprule
		& Cora & Citeseer & PubMed & Coauthor CS & Amazon Photos \\
		\midrule
		MLP& 55.1$\pm$1.4 & 59.1$\pm$1.2 & 71.4$\pm$0.8 & 88.3$\pm$0.7 & 69.6$\pm$3.8 \\
		MoNet& 81.7$\pm$0.4 & 71.2$\pm$0.7 & 78.6$\pm0.5$ & 90.8$\pm$0.6 & 91.2$\pm$1.3 \\
		GCN& 81.5$\pm$0.5 & 70.9$\pm$0.5 & 79.0$\pm$0.3 & 91.1$\pm$0.5 & 91.2$\pm$1.2 \\
		GraphSAGE& 79.2$\pm$7.7 &71.6$\pm$1.9 & 77.4$\pm$2.2 &91.3$\pm$2.8 & 91.4$\pm$1.4\\
		GAT& 83.0$\pm$0.7 & 72.5$\pm$0.7 & 79.0$\pm$0.3 & 90.5$\pm$0.6 & 85.1$\pm$2.3 \\
		JKNet& 83.7$\pm$0.7 & 72.5$\pm$0.4 & 82.6$\pm$0.5 & 91.1$\pm$0.3 & 86.1$\pm$1.1 \\
		APPNP& 83.5$\pm$0.7 & \second{75.9$\pm$0.6} & 80.2$\pm$0.3 & 91.5$\pm$0.1 & 87.0$\pm$0.9 \\
		GPR-GNN& 83.8$\pm$0.9 & \second{75.9$\pm$0.7} & 82.3$\pm$0.2 & 91.8$\pm$0.1 & 87.0$\pm$0.9 \\
		CurvGN & 82.6$\pm$0.6 & 71.5$\pm$0.8 & 78.8$\pm$0.6 & 92.9$\pm$0.4 & 92.5$\pm$0.5 \\
		UFGConv\_S & 83.0$\pm$0.5 & 71.0$\pm$0.6 & 79.4$\pm$0.4 & 92.1$\pm$0.2 & 92.1$\pm$0.5 \\
		UFGConv\_R & 83.6$\pm$0.6 & \third{72.7$\pm$0.6} & 79.6$\pm$0.4 & 93.0$\pm$0.7 & 92.5$\pm$0.2 \\
		RC-UFG (Hom)& 84.4$\pm$0.7 & 72.5$\pm$0.7 & 82.9$\pm$0.2 & \first{94.2$\pm$0.9} & \third{93.5$\pm$0.7} \\
		RC-UFG (Het)& 80.6$\pm$0.4 & 71.7$\pm$0.6 & 79.6$\pm$0.4 & 90.4$\pm$1.2 & 89.5$\pm$1.9 \\
		AFRC & \second{85.3$\pm$1.2} & 71.7$\pm$1.1 & \second{85.9$\pm$0.3} & 92.8$\pm$0.7 & 90.2$\pm$2.9 \\
		ARGNN & \third{85.2$\pm$0.8} & 71.9$\pm$1.5 & 84.5$\pm$6.8 & \second{93.8$\pm$0.2} & 77.6$\pm$5.9 \\
		BORF & 77.7$\pm$1.7 & 67.8$\pm$0.3 & 85.3$\pm$0.3 & 93.0$\pm$0.5 & 85.7$\pm$4.3 \\
		SDRF & \second{85.3$\pm$1.2} & 71.6$\pm$1.2 & \third{85.8$\pm$0.1} & 92.1$\pm$0.2 & \second{94.0$\pm$0.5} \\
		\midrule
		GEGCN & \first{86.7$\pm$1.2} & \first{76.6$\pm$1.1} & \first{87.4$\pm$0.4} & \third{93.2$\pm$0.4} & \first{94.1$\pm$0.5} \\
		\bottomrule
	\end{tabular}
\end{table*}

\subsection{Main Results}

\paragraph{\textbf{Node Classification on Homophilic Graphs.}}
As demonstrated in Table \ref{tab:result_realworld_dataasets_5}, GEGCN performs strongly across the five homophilic benchmarks, ranks first on Cora, Citeseer, PubMed, and Amazon Photos and third on Coauthor CS.

Compared to classic message-passing baselines such as GCN, GAT, and GraphSAGE, GEGCN exhibits significant performance gains. For instance, GEGCN exceeds GCN by 5.2 and 5.7 percentage points on Cora and Citeseer, respectively. This substantial improvement can be attributed to the learnable geometry-aware normalization matrix $\hat{A}^*$. By integrating the Ricci flow sequence via LSTM, our model effectively re-weights edges based on their geometric evolution, and provides geometry-aware information propagation.

GEGCN also remains competitive with recent geometric methods including CurvGN, UFGConv, RC-UFG, AFRC, ARGNN, BORF, and SDRF. On PubMed, GEGCN obtains 87.4\%, compared with 85.9\% for the next-highest method, AFRC. GEGCN models the entire trajectory of structural evolution through $\mathcal{K}_{e}$ and $\mathcal{W}_{e}$. The ablations and flow-time analyses below show that encoding this trajectory is more effective than using individual states or simple aggregation rules.

\paragraph{\textbf{Node Classification on Traditional Heterophilic Graphs.}}

Table~\ref{tab:result_heterophilic_dataasets_5} presents the node classification results on five widely adopted heterophilic benchmarks. GEGCN ranks first on Chameleon and Actor, third on Cornell and Texas, and fourth on Wisconsin.

Since the WebKB graphs (Cornell, Texas, and Wisconsin) contain only 183--251 nodes, a single test fold comprises merely a few dozen nodes. The reported accuracies for competing geometric methods (AFRC, ARGNN, BORF, and GNRF) exhibit large standard deviations (e.g., $\pm$ 5.44 for ARGNN and $\pm$ 5.78 for BORF on Wisconsin). Consequently, rankings on the WebKB graphs are sensitive to random splits and cannot be regarded as a definitive ordering of methods. Instead of claiming uniform superiority across every individual benchmark, GEGCN achieves the highest accuracy on the two larger heterophilic graphs (Chameleon and Actor) and yields substantially smaller standard deviations, demonstrating more stable performance relative to competing approaches.

It is notable that the methods relying on global structural modifications, such as diffusion-based preprocessing and curvature-guided rewiring, exhibit dataset-dependent behavior. While these approaches can partially alleviate connectivity sparsity, their performance varies considerably across benchmarks. For instance, DIGL offers limited or inconsistent improvements on Texas and Wisconsin (Table \ref{tab:result_heterophilic_dataasets_5}), despite enhancing graph connectivity. This suggests that, in heterophilic graphs, increasing connectivity alone is insufficient and may even be detrimental when newly introduced paths predominantly link nodes with dissimilar labels.

In contrast, GEGCN preserves the original edge support and assigns trajectory-conditioned propagation weights. Since the geometric states are derived from topology and edge weights rather than labels, this design does not assume that adjacent labels agree. We support these claims via empirical results, and do not guarantee that every learned weight yields benefits under heterophily.

\begin{table*}[htbp]
	\centering
	\caption{Mean test accuracy (\%) $\pm$ standard deviation across five traditional heterophilic benchmarks.}
	\label{tab:result_heterophilic_dataasets_5} 
	\small
	\begin{tabular}{lccccc} 
		\toprule
		& Cornell & Texas & Wisconsin & Chameleon & Actor \\
		\midrule
		GCN              & 52.69$\pm$0.21 & 61.19$\pm$0.49 & 54.60$\pm$0.86 & 41.33$\pm$0.18 & 23.84$\pm$0.43 \\
		Undirected        & 53.20$\pm$0.53 & 63.38$\pm$0.87 & 51.37$\pm$1.15 & 42.02$\pm$0.30 & 21.45$\pm$0.47 \\
		+FA               & 58.29$\pm$0.49 & 64.82$\pm$0.29 & 55.48$\pm$0.62 & 42.67$\pm$0.17 & 24.14$\pm$0.43 \\
		DIGL        & 58.26$\pm$0.50 & 62.03$\pm$0.43 & 49.53$\pm$0.27 & 42.02$\pm$0.13 & 24.77$\pm$0.32 \\
		DIGL + Undirected & 59.54$\pm$0.64 & 63.54$\pm$0.38 & 52.23$\pm$0.54 & 42.68$\pm$0.12 & 25.45$\pm$0.30 \\
		SDRF              & 54.60$\pm$0.39 & 64.46$\pm$0.38 & 55.51$\pm$0.27 & {42.73$\pm$0.15} & 28.42$\pm$0.75 \\
		SDRF + Undirected & 57.54$\pm$0.34 & 67.02$\pm$0.40 & 56.55$\pm$0.86 & 44.46$\pm$0.17 & {28.35$\pm$0.06} \\
		AFRC & 60.00$\pm$5.54 & 67.68$\pm$9.56 & 65.20$\pm$11.13 & \third{57.73$\pm$2.42} & 35.24$\pm$0.68 \\
		ARGNN & \second{72.59$\pm$4.52} & \second{77.16$\pm$4.52} & \third{73.04$\pm$5.44} & \second{58.92$\pm$3.23} & 32.86$\pm$1.27 \\
		BORF & 67.11$\pm$4.01 & 68.05$\pm$9.09 & \second{75.00$\pm$5.78} & 53.73$\pm$2.86 & \second{36.24$\pm$1.75} \\
		GNRF              & \first{73.05$\pm$3.52} & \first{83.33$\pm$6.08} & \first{88.20$\pm$3.84} & 53.82$\pm$3.02 & \third{35.53$\pm$0.68} \\
		\midrule
		GEGCN             & \third{68.61$\pm$0.26} & \third{70.27$\pm$0.69} & 67.39$\pm$0.93 & \first{60.72$\pm$0.16} & \first{37.18$\pm$0.30} \\
		\bottomrule
	\end{tabular}
\end{table*}

\paragraph{\textbf{Node Classification on Filtered Heterophilic Graphs.}}

Notably, recent work \citep{platonov2023criticallook} has identified train-test data leakage issues in the aforementioned heterophilic datasets, which may lead to inflated performance estimates and unfair benchmark comparisons. To evaluate GEGCN under a stricter protocol, we conduct additional experiments on the six heterophilic benchmarks without the train-test data leakage issue \citep{platonov2023criticallook}. As shown in Table \ref{tab:result_heterophilic_dataasets_modern}, GEGCN obtains the highest accuracy on Roman-empire, Amazon-ratings, and Minesweeper, ranks second on Questions and Squirrel-f, and ranks third on Chameleon-f. GEGCN thus ranks among the top methods on every filtered benchmark, confirming that its advantage carries over to the leakage-free setting. It is worth highlighting that GNRF, which delivers competitive performance on the unfiltered datasets (Table~\ref{tab:result_heterophilic_dataasets_5}), exhibits notably degraded performance on these rigorously filtered benchmarks.

\begin{table*}[htbp]
	\centering
	\caption{Mean test accuracy (\%) $\pm$ standard deviation across filtered benchmarks.}
	\label{tab:result_heterophilic_dataasets_modern} 
	\small
	\setlength{\tabcolsep}{4pt}
	\begin{tabular}{lcccccc} 
		\toprule
		& Roman-empire & Amazon-ratings & Minesweeper & Questions & Squirrel-f & Chameleon-f \\
		\midrule
		ResNet            & 65.88$\pm$0.38 & 45.90$\pm$0.52 & 50.89$\pm$1.39 & 70.34$\pm$0.76 & 36.55$\pm$1.82 & 36.73$\pm$4.71 \\
		GAT               & \second{80.87$\pm$0.30} & \third{49.09$\pm$0.63} & \second{92.01$\pm$0.68} & \third{77.43$\pm$1.20} & 35.62$\pm$2.06 & 39.21$\pm$3.08 \\
		H2GCN             & 60.11$\pm$0.52 & 36.47$\pm$0.23 & 89.71$\pm$0.31 & 63.59$\pm$1.46 & 35.10$\pm$1.15 & 26.75$\pm$3.64 \\
		GPR-GNN           & 64.85$\pm$0.27 & 44.88$\pm$0.34 & 86.24$\pm$0.61 & 55.48$\pm$0.91 & 35.11$\pm$1.24 & 25.90$\pm$3.58 \\
		GloGNN            & 59.63$\pm$0.69 & 36.89$\pm$0.14 & 51.08$\pm$1.23 & 65.74$\pm$1.19 & 35.51$\pm$1.65 & 39.61$\pm$2.60 \\
		FSGNN             & \third{79.92$\pm$0.56} & \second{52.74$\pm$0.83} & 90.08$\pm$0.70 & \first{78.86$\pm$0.92} & 35.92$\pm$1.32 & \second{40.61$\pm$2.97} \\
		FAGCN             & 65.22$\pm$0.56 & 44.12$\pm$0.30 & 88.17$\pm$0.73 & 77.24$\pm$1.26 & \third{38.95$\pm$1.99} & 39.93$\pm$3.30 \\
		GBK-GNN           & 74.57$\pm$0.47 & 45.98$\pm$0.71 & \third{90.85$\pm$0.58} & 74.47$\pm$0.86 & \first{41.08$\pm$2.27} & \first{41.90$\pm$2.72} \\
		GNRF              & 79.72$\pm$0.77 & 40.04$\pm$0.95 & 84.81$\pm$1.25 & 70.38$\pm$3.75 & 31.35$\pm$1.79 & 34.77$\pm$2.33 \\
		\midrule
		GEGCN             & \first{86.74$\pm$1.02} & \first{52.77$\pm$0.73} & \first{92.25$\pm$0.93} & \second{77.73$\pm$0.41} & \second{40.03$\pm$1.71} & \third{40.53$\pm$1.81} \\
		\bottomrule
	\end{tabular}
\end{table*}

Across the five traditional and six filtered heterophilic datasets, GEGCN obtains five first-place, two second-place, three third-place, and one fourth-place results, yielding an average rank of 2.00. Such consistency and stability across all eleven heterophilic benchmarks constitute the main evidence supporting GEGCN's advantages under heterophilic settings.

\paragraph{\textbf{Large-Scale Evaluation.}}

To evaluate the large-scale performance and computational efficiency of GEGCN, we also conduct experiments on the OGBN-arXiv benchmark,  which is a typical large-scale sparse graph with complex topology. On OGBN-arXiv, we compare GEGCN with representative baselines covering multiple mainstream categories. Experimental results (Table \ref{tab:result_heterophilic_dataset_large}) demonstrate that GEGCN achieves the highest accuracy among the compared methods on this large-scale benchmark.

The performance of GEGCN indicates that the geometric evolution combined with LSTM remains effective on large-scale graphs. The result on OGBN-arXiv complements the small- and medium-scale benchmarks and verifies GEGCN's competitiveness across varying graph sizes. We further report the training and inference time of GEGCN on 7 datasets in Section \ref{sec:complexity}, together with its offline Ricci-flow preprocessing time over all 17 datasets in Appendix~\ref{time}, demonstrating that GEGCN maintains acceptable computational overhead while achieving competitive accuracy.

Overall, GEGCN achieves an average rank of 1.76 across all 17 datasets. Such consistent performance across diverse settings demonstrates that GEGCN exhibits strong generalization and adaptability to various graph types.

\begin{table}[htbp]
	\centering
	\caption{Mean test accuracy (\%) $\pm$ standard deviation on OGBN-arXiv.}
	\label{tab:result_heterophilic_dataset_large} 
	\small
	\begin{tabular}{lc} 
		\toprule
		& Test Accuracy \\
		\midrule
		MLP       & 57.65$\pm$0.12 \\
		Node2vec  & 71.29$\pm$0.13 \\
		GraphSAGE & 72.77$\pm$0.16 \\
		GCN       & 73.00$\pm$0.17 \\
		GAT       & 73.30$\pm$0.16 \\
		UniMP     & 74.50$\pm$0.05 \\
		BiGTex & \third{74.62$\pm$2.16} \\
		AGDN      & \second{75.22$\pm$0.07} \\
		\midrule
		GEGCN     & \first{75.76$\pm$0.64} \\
		\bottomrule
	\end{tabular}
\end{table}

\subsection{Ablation Studies}

We conduct extensive ablation experiments to verify the contribution of components and design choices in GEGCN. Specifically, these experiments include comparisons among LSTM encoder and sequence aggregation variants, order perturbation on the geometric states generated by Ricci flow, a propagation-only diagnostic for the geometry-aware matrix, as well as depth stress tests and flow-duration analysis. All entries report the mean test accuracy over 10 runs.

The first state and last state variants test individual Ricci-flow states. Mean pooling and max pooling test non-recurrent trajectory aggregation. The MLP variant tests a non-recurrent learnable encoder. The shuffled and reversed variants test whether retaining the same states without their forward order is sufficient. The propagation-only comparison between $\hat{\mathbf{A}}$ and $\hat{\mathbf{A}}^*$ isolates the contribution of the geometry-aware propagation matrix, while the depth stress test and flow-duration analysis evaluated deep propagation stability and the contribution of trajectory length, respectively.

\paragraph{\textbf{Encoder and Aggregation Variants.}}
We compare the LSTM encoder with multiple alternative aggregation variants on Actor, Cornell, and Wisconsin. First, we consider a baseline using standard GCN, where both the Ricci flow evolution module and the LSTM encoder are removed. Second, we retain the Ricci flow module and replace the LSTM with various pooling strategies to model curvature sequences, including selecting the first or last state, mean pooling, and max pooling. Third, we maintain the GEGCN pipeline and simply replace the LSTM encoder with an MLP. Since this MLP variant retains the overall GEGCN framework and only substitutes the LSTM encoder, it is different from GNRF, which builds upon the diffusion equation coupled with an MLP-based EdgeNet.

Panel A of Table~\ref{tab:ablation_studies} lists these ablation configurations and experimental results, together with the average accuracy gaps compared with the full GEGCN results in Table~\ref{tab:result_heterophilic_dataasets_5}. The full GEGCN model obtains the highest mean accuracy on Actor, Cornell, and Wisconsin. Among non-recurrent sequence aggregation strategies, mean pooling, first-state encoding, and last-state encoding have average changes of $-3.9$, $-4.0$, and $-4.5$ percentage points, respectively, relative to GEGCN. The larger decreases for max pooling and the MLP variant ($-17.5$ and $-20.2$ percentage points) indicate that simply retaining the Ricci flow evolution is not sufficient to match the full GEGCN with its LSTM encoder. In this comparison, LSTM-based trajectory encoding outperforms the evaluated static and order-insensitive alternatives.

\begin{table*}[htbp]
	\centering
	\caption{\textbf{Ablations on Encoder Variants and Sequence Ordering.} Mean test accuracy (\%) over 10 runs. For each panel, Avg.~$\Delta$ is the average variant accuracy minus the corresponding GEGCN accuracy; negative values indicate a decrease.}
	\label{tab:ablation_studies}
	\small
	\setlength{\tabcolsep}{4pt}
	\begin{tabular}{@{}lccrrrr@{}}
		\toprule
		\multicolumn{7}{l}{\textbf{Panel A: Ablations on Encoder Variants}} \\
		\midrule
		\textbf{Variant} & \shortstack{\textbf{Ricci}\\\textbf{flow}} & \shortstack{\textbf{Temporal}\\\textbf{encoder}} & \textbf{Actor} & \textbf{Cornell} & \textbf{Wisconsin} & \textbf{Avg. $\Delta$} \\
		\midrule
		GCN & $\times$ & -- & 23.8 & 52.1 & 54.6 & -14.2 \\
		First state & $\checkmark$ & First & 33.6 & 63.6 & 64.0 & -4.0 \\
		Last state & $\checkmark$ & Last & 33.9 & 61.4 & 64.4 & -4.5 \\
		Mean pooling & $\checkmark$ & Mean & 34.4 & 62.8 & 64.4 & -3.9 \\
		Max pooling & $\checkmark$ & Max & 29.1 & 41.7 & 50.0 & -17.5 \\
		MLP & $\checkmark$ & Order-insensitive & 26.2 & 41.7 & 44.8 & -20.2 \\
		\textbf{GEGCN} & $\checkmark$ & LSTM & \textbf{37.2} & \textbf{68.6} & \textbf{67.4} & -- \\
		\midrule
		\multicolumn{7}{l}{\textbf{Panel B: Ablations on Sequence Ordering}} \\
		\midrule
		\textbf{Variant} & \shortstack{\textbf{Ricci}\\\textbf{flow}} & \shortstack{\textbf{Sequence}\\\textbf{order}} & \textbf{Cora} & \textbf{Citeseer} & \textbf{Actor} & \textbf{Avg. $\Delta$} \\
		\midrule
		Shuffled Ricci & $\checkmark$ & Shuffled & 81.9 & 70.2 & 31.1 & -5.8 \\
		Reversed Ricci & $\checkmark$ & Reversed & 84.8 & 72.9 & 34.7 & -2.7 \\
		\textbf{Forward Ricci (GEGCN)} & $\checkmark$ & Forward & \textbf{86.7} & \textbf{76.6} & \textbf{37.2} & -- \\
		\bottomrule
	\end{tabular}
\end{table*}

\paragraph{\textbf{Order Perturbations.}}
To verify whether the temporal ordering of geometric states along Ricci flow carries meaningful information, we conduct ordering perturbations on the geometric state sequence fed into the encoder, including random shuffling and sequence reversal. These two variants preserve the same geometric states as GEGCN but alter their order, thereby isolating sensitivity to the forward flow-time sequence.

We perform this set of experiments on Cora, Citeseer, and Actor and report the results in Panel B of Table~\ref{tab:ablation_studies}. Both perturbations reduce mean accuracy on all evaluated datasets. Reversed order yields an average change of $-2.7$ percentage points, while random shuffling yields a larger average change of $-5.8$ percentage points. The consistent decreases indicate that the forward Ricci-flow ordering contributes beyond mere access to the same set of states.

Taken together, Panels A and B highlight the joint contribution of Ricci-flow geometric states and sequential encoding by the LSTM. Static aggregation cannot adequately model geometric evolution, and perturbing the order of geometric states impairs performance substantially. These observations support that the LSTM captures evolving geometric patterns when provided with the properly ordered sequence of geometric states.

\paragraph{\textbf{Analysis of Over-Smoothing.}}
Deep message passing tends to render node representations progressively indistinguishable as more propagation layers are stacked. We investigate this phenomenon from two perspectives. First, we conduct a depth robustness stress test using two different depth configurations (2 and 32 layers) and compare the resulting node classification performance. Second, we further conduct a propagation-only diagnostic to evaluate the representation diversity induced by the propagation matrix at depths \(\ell\in\{1,2,4,8,16,32\}\).

We adopt four metrics to quantify over-smoothing. The accuracy retention ratio is defined as \(\text{Acc@32} / \text{Acc@2}\), which quantifies the fraction of predictive accuracy preserved when scaling the network depth from 2 layers to 32 layers. The Dirichlet energy \(\mathcal{E}(\mathbf{H})=\frac{1}{2}\sum_{(i,j)\in\mathcal{E}}\|\mathbf{h}_i-\mathbf{h}_j\|_2^2\) captures the variation of node representations across graph edges. The average cosine distance quantifies pairwise feature separation. The cosine distance is defined as \(\boldsymbol{d}_{\mathrm{cos}}(\boldsymbol{h}_i,\boldsymbol{h}_j) = 1 - \cos(\boldsymbol{h}_i,\boldsymbol{h}_j)\). The effective rank \(\mathrm{erank}(\mathbf{H})=\exp(-\sum_k p_k\log p_k)\), where \(p_k=\sigma_k/\sum_j\sigma_j\) and \(\{\sigma_k\}\) denote the singular values of \(\mathbf{H}\), measures the dimensional diversity of the representation matrix. Larger values of these metrics indicate that propagated features maintain greater variation and suffer less degeneracy.

Table~\ref{tab:depth_robustness} presents model performance at 2 and 32 layers, alongside the accuracy retention ratio \(\text{Acc@32} / \text{Acc@2}\), the average cosine distance and effective rank evaluated at depth 32. Across the tested datasets, GEGCN preserves substantially higher accuracy and representation diversity at large depth, indicating that the geometric propagation operator delays depth-induced representation collapse.

\begin{table*}[htbp]
	\centering
	\caption{Depth-robustness stress test. Ret. denotes Acc@32/Acc@2; Cos.@32 and Rank@32 are average cosine distance and effective rank at depth 32.}
	\label{tab:depth_robustness}
	\small
	\setlength{\tabcolsep}{6pt}
	\begin{tabular}{llccccc}
		\toprule
		Dataset & Model & Acc@2 & Acc@32 & Ret. & Cos.@32 & Rank@32 \\
		\midrule
		\multirow{3}{*}{Actor}
		& GCN   & 23.84 & 19.20 & 80.5\% & 0.0064 & 2.05 \\
		& GAT   & 28.05 & 22.93 & 81.7\% & 0.0374 & 9.26 \\
		& GEGCN & \textbf{37.18} & \textbf{32.47} & \textbf{87.3\%} & \textbf{0.2483} & \textbf{17.62} \\
		\midrule
		\multirow{3}{*}{Citeseer}
		& GCN   & 70.90 & 22.21 & 31.3\% & 0.0021 & 1.75 \\
		& GAT   & 72.50 & 22.15 & 30.6\% & 0.0115 & 6.27 \\
		& GEGCN & \textbf{76.60} & \textbf{65.83} & \textbf{85.9\%} & \textbf{0.2214} & \textbf{15.74} \\
		\midrule
		\multirow{3}{*}{PubMed}
		& GCN   & 79.00 & 39.50 & 50.0\% & 0.0180 & 2.40 \\
		& GAT   & 79.00 & 37.80 & 47.8\% & 0.0200 & 6.80 \\
		& GEGCN & \textbf{87.40} & \textbf{78.82} & \textbf{90.2\%} & \textbf{0.2361} & \textbf{16.88} \\
		\bottomrule
	\end{tabular}
\end{table*}

We conduct the propagation-only diagnostic on Actor, Citeseer, and Cora. This diagnostic removes the classifier, nonlinear activation functions, and trainable feature transformations. It directly compares repeated propagation by the standard normalized adjacency matrix \(\hat{\mathbf{A}}\) in Eq.~\eqref{adjacencymatrix} and by the geometry-aware matrix \(\hat{\mathbf{A}}^*\) defined in the curvature-aware graph convolution module. Starting from the same input feature matrix \(\mathbf{X}\), we compute \(\mathbf{H}_{A}^{(0)}=\mathbf{X}\) and \(\mathbf{H}_{A}^{(\ell)}=A\mathbf{H}_{A}^{(\ell-1)}\) for \(A\in\{\hat{\mathbf{A}},\hat{\mathbf{A}}^*\}\), and record the propagated representations at depths \(\ell\in\{1,2,4,8,16,32\}\).

Figure~\ref{fig:oversmoothing_metrics} shows that \(\hat{\mathbf{A}}^*\) consistently preserves more representation diversity than \(\hat{\mathbf{A}}\) as the propagation depth increases. The advantage is especially clear at 16 and 32 layers, where \(\hat{\mathbf{A}}^*\) maintains larger cosine distance and effective rank across all three datasets. Its Dirichlet energy also remains higher in the deep propagation regime, showing that the node features collapse less severely into an overly smooth subspace. Since this diagnostic removes the classifier and other trainable transformations, the observed improvement can be attributed to the propagation matrix itself. These results support the claim that the geometry-aware propagation matrix mitigates representation collapse across the evaluated depths by preserving pairwise feature separation and the rank structure of node representations.

\begin{figure*}[htbp]
	\centering
	\begin{tikzpicture}
		\pgfplotstableread{
			depth anorm_dir astar_dir anorm_cos astar_cos anorm_rank astar_rank
			1 2027.60 2043.71 0.6709 0.6742 529.79 529.94
			2 1296.48 1358.42 0.5771 0.6083 438.98 487.63
			4 625.97 754.86 0.4982 0.5571 322.09 401.28
			8 477.04 611.93 0.4495 0.5188 199.00 308.55
			16 670.82 787.24 0.4498 0.5216 108.04 227.41
			32 1306.52 1487.65 0.5406 0.5783 58.15 158.72
		}\actordata
		\pgfplotstableread{
			depth anorm_dir astar_dir anorm_cos astar_cos anorm_rank astar_rank
			1 8.3329 8.4213 0.9019 0.9083 1572.27 1574.62
			2 3.4041 4.1832 0.8655 0.8924 1226.83 1357.41
			4 2.1422 3.2147 0.8219 0.8583 901.61 1093.28
			8 1.6090 2.5863 0.7695 0.8237 690.84 937.55
			16 1.3294 2.1094 0.7090 0.7772 566.66 753.19
			32 1.1803 1.6942 0.6470 0.7431 489.15 647.83
		}\citeseerdata
		\pgfplotstableread{
			depth anorm_dir astar_dir anorm_cos astar_cos anorm_rank astar_rank
			1 6.7003 6.8124 0.8497 0.8583 915.23 919.37
			2 2.7499 4.1738 0.7604 0.8124 739.79 806.42
			4 2.2315 3.6215 0.6532 0.7372 510.18 673.55
			8 1.9395 2.9863 0.5221 0.6637 317.44 547.28
			16 1.7269 2.5142 0.3832 0.5673 191.88 413.66
			32 1.5803 2.0917 0.2641 0.4831 122.70 317.94
		}\coradata
		\begin{groupplot}[
			group style={group size=3 by 3,horizontal sep=1.05cm,vertical sep=0.78cm},
			width=0.285\textwidth,
			height=0.175\textwidth,
			xmode=log,
			log basis x=2,
			xmin=0.9,
			xmax=36,
			xtick={1,2,4,8,16,32},
			xticklabels={1,2,4,8,16,32},
			tick align=outside,
			tick pos=left,
			axis x line*=bottom,
			axis y line*=left,
			tick label style={font=\scriptsize},
			label style={font=\scriptsize},
			title style={font=\small\bfseries},
			grid=major,
			grid style={draw=gray!18,line width=0.25pt},
			axis line style={gray!65},
			every axis plot/.append style={line width=1.15pt,mark size=1.7pt}
			]
			\nextgroupplot[
			title={Dirichlet Energy},ylabel={Actor},
			ylabel style={font=\small\bfseries,rotate=-90,at={(-0.30,0.5)}},
			ymin=300,ymax=2200,ytick={500,1000,1500,2000},xticklabels=\empty,
			legend to name=oversmoothinglegend,
			legend style={draw=none,font=\small,legend columns=2,
				/tikz/every even column/.append style={column sep=0.45cm}}
			]
			\addplot+[AnormColor,mark=*] table[x=depth,y=anorm_dir]{\actordata};
			\addlegendentry{$\hat{\mathbf{A}}$}
			\addplot+[AstarColor,mark=square*] table[x=depth,y=astar_dir]{\actordata};
			\addlegendentry{$\hat{\mathbf{A}}^*$}
			\nextgroupplot[title={Cosine Distance},ymin=0.42,ymax=0.70,
			ytick={0.45,0.55,0.65},xticklabels=\empty]
			\addplot+[AnormColor,mark=*] table[x=depth,y=anorm_cos]{\actordata};
			\addplot+[AstarColor,mark=square*] table[x=depth,y=astar_cos]{\actordata};
			\nextgroupplot[title={Effective Rank},ymin=0,ymax=560,
			ytick={0,200,400},xticklabels=\empty]
			\addplot+[AnormColor,mark=*] table[x=depth,y=anorm_rank]{\actordata};
			\addplot+[AstarColor,mark=square*] table[x=depth,y=astar_rank]{\actordata};
			\nextgroupplot[ylabel={Citeseer},
			ylabel style={font=\small\bfseries,rotate=-90,at={(-0.30,0.5)}},
			ymin=0.6,ymax=8.8,ytick={2,4,6,8},xticklabels=\empty]
			\addplot+[AnormColor,mark=*] table[x=depth,y=anorm_dir]{\citeseerdata};
			\addplot+[AstarColor,mark=square*] table[x=depth,y=astar_dir]{\citeseerdata};
			\nextgroupplot[ymin=0.62,ymax=0.93,ytick={0.65,0.75,0.85},xticklabels=\empty]
			\addplot+[AnormColor,mark=*] table[x=depth,y=anorm_cos]{\citeseerdata};
			\addplot+[AstarColor,mark=square*] table[x=depth,y=astar_cos]{\citeseerdata};
			\nextgroupplot[ymin=400,ymax=1650,ytick={500,1000,1500},xticklabels=\empty]
			\addplot+[AnormColor,mark=*] table[x=depth,y=anorm_rank]{\citeseerdata};
			\addplot+[AstarColor,mark=square*] table[x=depth,y=astar_rank]{\citeseerdata};
			\nextgroupplot[xlabel={Depth},ylabel={Cora},
			ylabel style={font=\small\bfseries,rotate=-90,at={(-0.30,0.5)}},
			ymin=1.0,ymax=7.2,ytick={2,4,6}]
			\addplot+[AnormColor,mark=*] table[x=depth,y=anorm_dir]{\coradata};
			\addplot+[AstarColor,mark=square*] table[x=depth,y=astar_dir]{\coradata};
			\nextgroupplot[xlabel={Depth},ymin=0.22,ymax=0.90,ytick={0.3,0.5,0.7,0.9}]
			\addplot+[AnormColor,mark=*] table[x=depth,y=anorm_cos]{\coradata};
			\addplot+[AstarColor,mark=square*] table[x=depth,y=astar_cos]{\coradata};
			\nextgroupplot[xlabel={Depth},ymin=80,ymax=980,ytick={200,500,800}]
			\addplot+[AnormColor,mark=*] table[x=depth,y=anorm_rank]{\coradata};
			\addplot+[AstarColor,mark=square*] table[x=depth,y=astar_rank]{\coradata};
		\end{groupplot}
		\node at ($(group c2r1.north)+(0,0.90cm)$)
		{\pgfplotslegendfromname{oversmoothinglegend}};
	\end{tikzpicture}
	\caption{\textbf{Propagation-only diagnostic.} Repeated propagation by the standard normalized adjacency and the learned geometry-aware matrix is compared on Actor, Citeseer, and Cora at increasing depths. Higher Dirichlet energy, average cosine distance, and effective rank indicate greater retained representation variation.}
	\label{fig:oversmoothing_metrics}
\end{figure*}

\paragraph{\textbf{Impact of Ricci Flow Duration.}}
\begin{figure}[htbp]
	\centering
	\includegraphics[trim={0mm 0mm 0mm 9.5mm}, clip, width=\linewidth]{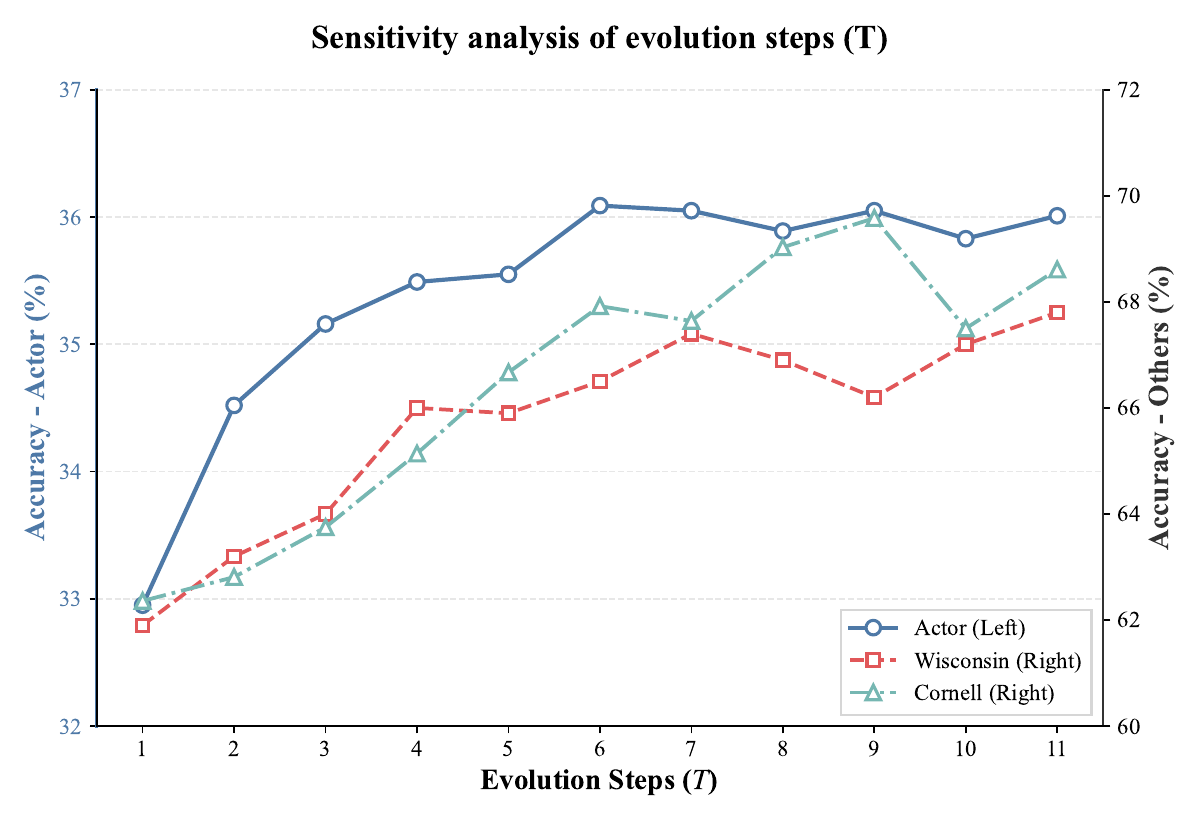}
	\caption{\textbf{Sensitivity to evolution steps $T$.} Accuracy generally improves as additional Ricci-flow states are included up to approximately six steps and then tends to saturate.}
	\label{fig:Sensitivity}
\end{figure}

Figure~\ref{fig:Sensitivity} reports the impact of the discrete Ricci flow step count $T$ on model performance. We observe a consistent accuracy gain as $T$ increases from 1 to 6 across three datasets. This trend shows that GEGCN benefits from integrating a longer sequence of Ricci-flow geometric states instead of relying on a single static graph snapshot. Specifically, the Actor dataset exhibits the most significant improvement, likely due to its highly sparse nature, which requires more evolution steps to reveal underlying homophily clusters.

\paragraph{\textbf{Distinctness of Ricci-Flow Geometric States.}}
The multi-scale interpretation requires the sampled flow times to provide different geometric states. For $s,t\in\{0,\ldots,T\}$, we compare curvature vectors under a fixed edge ordering using
\begin{gather*}
D(s,t)=\frac{\lVert\bm\kappa(s)-\bm\kappa(t)\rVert_2}{\sqrt{|\mathcal E|}},\\
\rho_{\mathrm S}(s,t)=\operatorname{Corr}(\mathbf r(s),\mathbf r(t)),
\end{gather*}
where $\mathbf r(t)=\operatorname{rank}(\boldsymbol{\kappa}(t))$ is the edgewise rank vector.

\begin{figure*}[htbp]
	\centering
	\includegraphics[width=0.85\textwidth]{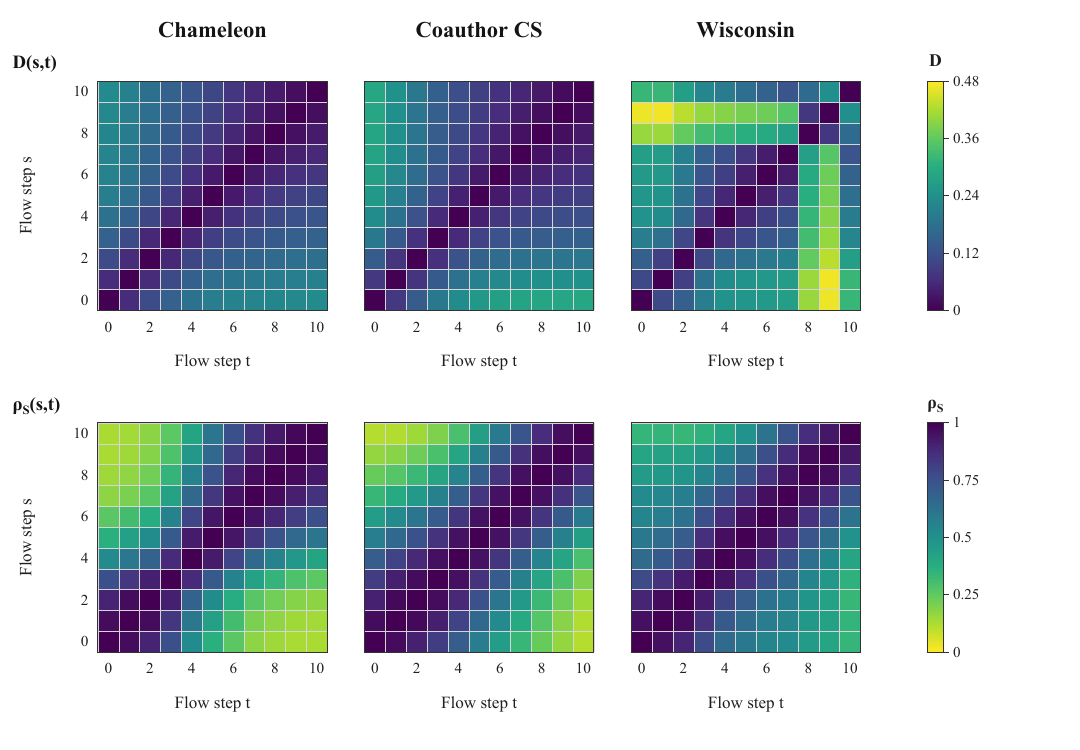}
	\caption{\textbf{Pairwise distinctness of flow-time states.} Curvature-vector distances $D(s,t)$ (top) and Spearman correlations $\rho_{\mathrm S}(s,t)$ (bottom) for 11 states sampled at $t=0,\ldots,10$ on Chameleon, Coauthor CS, and Wisconsin.}
	\label{fig:flow_time_state_distinctness}
\end{figure*}

Across all three datasets, every off-diagonal $D(s,t)$ in Figure~\ref{fig:flow_time_state_distinctness} is positive. The smallest values between distinct times are $0.017$ (Chameleon), $0.015$ (Coauthor CS), and $0.037$ (Wisconsin). No two sampled times therefore share the same curvature vector, so the states $\mathcal{S}_t$ are pairwise distinct. The change also grows with the time gap. The endpoint distance $D(0,10)$ reaches $0.227$, $0.286$, and $0.319$, while the Spearman correlation between the first and last states drops to $0.127$, $0.108$, and $0.341$. The low correlations show that the flow reorders which edges are most curved, not merely rescales the curvature magnitudes.

\subsection{Computational Complexity}\label{sec:complexity}

The geometric evolution process comprises shortest-path recomputation and edgewise optimal transport. For an edge $(u,v)$, each Sinkhorn iteration~\citep{cuturi2013sinkhorn} costs $O(d_u d_v)$ once the transport-cost matrix is available, where $d_u$ and $d_v$ denote the degrees of nodes $u$ and $v$, respectively. With at most $K$ Sinkhorn iterations per edge, the entropy-regularized transport computations cost $O(KQ)$ per flow state, where $Q=\sum_{(u,v)\in\mathcal E}d_u d_v$. Shortest-path distances are recomputed after each simultaneous update of the edge weights. With a Fibonacci-heap implementation of all-sources Dijkstra's algorithm, this requires $O(|\mathcal V|(|\mathcal E|+|\mathcal V|\log|\mathcal V|))$ per flow step. Treating the solver iteration bound $K$ and the number of Ricci-flow updates $T$ as constants independent of graph size, the total preprocessing cost is
\[
O\!\left(
|\mathcal V|
\bigl(|\mathcal E|+|\mathcal V|\log|\mathcal V|\bigr)+Q
\right).
\]

The dominant term in this bound depends on graph size, sparsity, and degree distribution. If node degrees are comparable to the average degree $D=2|\mathcal E|/|\mathcal V|$, then $Q=O(|\mathcal E|D^2)$, giving the bound $O(|\mathcal V||\mathcal E|+|\mathcal V|^2\log|\mathcal V|+|\mathcal E|D^2)$. For graphs with uniformly bounded node degrees, $|\mathcal E|=O(|\mathcal V|)$ and $Q=O(|\mathcal V|)$, so the $O(|\mathcal V|^2\log|\mathcal V|)$ shortest-path term dominates the stated bound. For dense graphs with $|\mathcal E|=\Theta(|\mathcal V|^2)$ and node degrees $\Theta(|\mathcal V|)$, the transport and shortest-path terms are $O(|\mathcal V|^4)$ and $O(|\mathcal V|^3)$, respectively, making transport the dominant term in the bound.

The discrete Ricci flow is implemented as a one-time preprocessing step (see the execution time in Appendix \ref{time}), which is decoupled from the neural network training. Consequently, the training phase involves only the LSTM encoder and GCN layers, both of which are linear with respect to the number of edges, ensuring the overall GEGCN framework remains efficient and feasible for practical applications.

\begin{table}[htbp]
	\centering
	\caption{The average training and inference time of GEGCN across seven datasets.}
	\label{tab:time} 
	\small
	\begin{tabular}{lcc} 
		\toprule
		&Train (s) & Inf. (s) \\
		\midrule
		Chameleon-f    & 34.62  & 0.078 \\
		Minesweeper    & 137.90 & 0.574 \\
		Squirrel-f     & 153.96 & 0.597 \\
		Roman-empire   & 158.84 & 0.353 \\
		Amazon-ratings & 244.22 & 1.357 \\
		Questions      & 457.05 & 1.759 \\
		OGBN-arXiv     & 732.90 & 14.557 \\		
		\bottomrule
	\end{tabular}
\end{table}

To further assess the computational efficiency of GEGCN, we report its average training and inference time on 7 datasets in Table~\ref{tab:time}, together with its offline Ricci-flow preprocessing time for all 17 datasets in Appendix~\ref{time}. Even on the large-scale OGBN-arXiv, the training and inference times remain manageable. The acceptable computational efficiency and stable performance on various graphs further illustrate that GEGCN has good practical deployment potential for real large-scale graphs.

\section{Conclusion}

The core idea is to encode the edgewise curvature--weight sequences generated by discrete Ricci flow and use the resulting scores to define a geometry-aware propagation matrix. GEGCN first computes multiple flow-time geometric states, then encodes each edge sequence with an LSTM, and finally maps the encoded sequence to an edge score used in the normalized graph-convolution operator. The encoded quantities are not appended as node features.

The core idea of this paper is to model the evolutionary sequences generated by Ricci flow as a temporal process and integrate them into a graph deep learning framework. Our framework consists of three stages: (i) We first apply discrete Ricci flow to the input graph to generate multi-scale geometric states. (ii) An LSTM encoder is then employed to process this sequence, learning to encode the ordered Ricci-flow states into a compact representation. (iii) Finally, these geometry-aware representations are mapped to edge scores and incorporated into a standard GCN, which performs the downstream prediction tasks. This design allows GEGCN to simultaneously leverage local neighborhood information (via GCN) and Ricci-flow geometric states (via Ricci flow and LSTM). Extensive experiments across 17 datasets demonstrate the competitive performance of the proposed method across diverse graph settings. More broadly, GEGCN offers a modular perspective on geometry-aware graph learning: rather than treating curvature as a static descriptor or relying solely on topology rewiring, it models geometric evolution as a learnable sequential signal and integrates the resulting structural dynamics directly into message passing. We hope this perspective will provide a principled bridge between discrete geometric analysis and neural graph representation learning, and inspire future graph architectures that more fully exploit geometric dynamics.


\section*{Data and code availability}

All datasets are publicly available from the sources cited in Section \ref{exp_data}. The source-code repository for GEGCN is available at \url{https://github.com/mjc191812/Geometric-Evolution-Graph-Convolutional-Networks}.

\section*{Acknowledgements}

The research is supported by the National Natural Science Foundation of China (No. 12271039).

\section*{Impact Statement}

This paper presents the Geometric Evolution Graph Convolutional Network (GEGCN), a novel framework that enhances graph representation learning by modeling geometric evolution on graphs. There are many potential societal consequences of our work, none of which we feel must be specifically highlighted here.

\bibliographystyle{unsrtnat}
\bibliography{GEGN}
\newpage
\appendix
\onecolumn
\counterwithin{table}{section}
\counterwithin{figure}{section}

\section{Discrete Ricci Curvature and Ricci Flow}
\label{app:riccicurvature}

\subsection{Ollivier-Ricci Curvature on Graphs}

Various notions of Ricci curvature have been proposed for graphs, with the aim of capturing their geometric and topological properties, including Forman-Ricci curvature~\citep{forman2003bochner}, Menger-Ricci curvature~\citep{Menger1930}, Haantjes-Ricci curvature~\citep{haantjes1947discrete}, and Bakry-\'Emery Ricci curvature~\citep{bakry1985hyper}. Among the various definitions proposed in the literature, Ollivier-Ricci curvature~\citep{ollivier2009ricci}, with its foundation in optimal transport theory and clear geometric meaning, has gained significant prominence. It quantifies curvature by measuring the transportation cost of probability measures centered at adjacent nodes. In this work, we focus on Ollivier-Ricci curvature as our primary notion of discrete Ricci curvature.

Let $\mathcal{G} = (\mathcal{V}, \mathcal{E})$ be a graph with node set $\mathcal{V} = \{v_1, \dots, v_N\}$ and edge set $\mathcal{E} \subseteq \mathcal{V} \times \mathcal{V}$. Each edge $e\in \mathcal{E}$ is assigned an edge weight $w_e$.  
The distance between two nodes $u$ and $v$ is
$\rho_{uv}=\inf_{\gamma}\sum_{e\in \gamma} w_e$, where $\gamma$ ranges over all paths connecting $u$ and $v$.
For a random walk with parameter $\alpha\in[0,1]$, the probability measure centered at $u\in \mathcal{V}$ is
\[
m_u^\alpha(v)=
\begin{cases}
	\alpha, & v=u,\\[0.5em]
	(1-\alpha)\dfrac{w_{uv}}{\sum_{z\sim u}w_{zu}}, & v\sim u,\\
	0, & \text{otherwise}.
\end{cases}
\]
For two probability measures $m^\alpha_u,m^\alpha_v$ on $\mathcal{V}$, a coupling $A:\mathcal{V}\times \mathcal{V}\to[0,1]$ satisfies
\[
\sum_{y\in \mathcal{V}} A(x,y)=m^\alpha_u(x), \quad
\sum_{x\in \mathcal{V}} A(x,y)=m^\alpha_v(y).
\]
The Wasserstein distance between $m^\alpha_u$ and $m^\alpha_v$ is
\[
W(m^\alpha_u,m^\alpha_v) = \inf_A \sum_{x,y\in \mathcal{V}} A(x,y)\rho_{xy}.
\]
Ollivier-Ricci curvature of edge $e=(u,v)$ is defined as
\[
\kappa_e^\alpha = 1 - \frac{W(m_u^\alpha,m_v^\alpha)}{\rho_{e}}.
\]

\begin{figure*}[h]
	\centering
	\includegraphics[width=12cm]{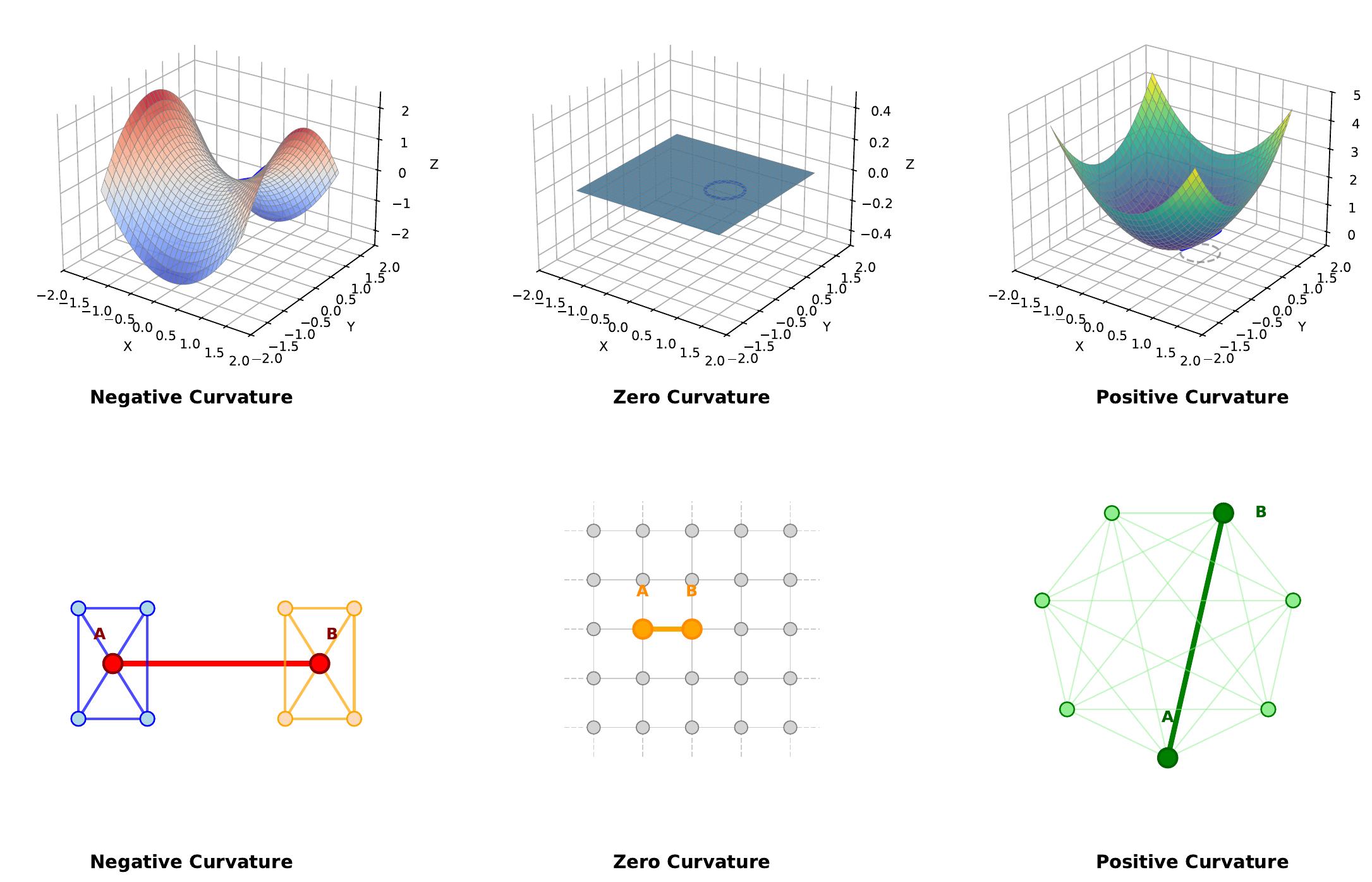}       
	\caption{\textbf{Comparison of Ollivier-Ricci curvature in continuous manifolds and discrete graphs.} 
		(Top row) Continuous manifolds: (Left) Hyperbolic paraboloid with negative curvature ($\kappa < 0$); 
		(Middle) Plane with zero curvature ($\kappa = 0$); (Right) Elliptic paraboloid with positive curvature ($\kappa > 0$). 
		(Bottom row) Discrete graphs: (Left) Two dense clusters connected by a single edge with negative curvature; 
		(Middle) Regular square lattice with zero curvature; (Right) Complete graph $K_7$ with positive curvature. 
		All edge weights are set to 1.}
	\label{fig1}
\end{figure*}

From a geometric perspective, positive Ollivier-Ricci curvature indicates a significant overlap between the neighborhoods of $u$ and $v$, implying a local tendency of information or mass to concentrate, analogous to the convergence of geodesics on positively curved manifolds. In contrast, negative curvature corresponds to a separation of neighborhoods and a rapid spreading of information, mirroring the exponential divergence of geodesics in negatively curved spaces. Zero curvature represents a balanced configuration in which neither contraction nor expansion dominates, akin to the flat geometry of Euclidean space. Figure \ref{fig1} illustrates these interpretations by comparing continuous manifolds and discrete graphs with negative, zero, and positive Ollivier-Ricci curvature. The top row presents representative surfaces from continuous geometry, while the bottom row shows graph structures exhibiting the corresponding curvature behaviors, thereby providing an intuitive correspondence between the continuous and discrete settings.

\subsection{Discrete Ricci Flow}

In the smooth setting, Ricci flow provides a fundamental geometric mechanism for evolving the Riemannian metric, which plays a central role in geometric analysis. \cite{ollivier2009ricci} first observed that a discrete analogue of the Ricci flow on manifolds can be formulated on weighted graphs,
\[
w_e^{\prime}(t) = -\kappa_e(t) \, w_e(t), \quad \forall e\in \mathcal{E},
\]
where $\kappa_e(t)$ denotes the Ollivier-Ricci curvature of edge $e$ at time $t$. Subsequently, \cite{bai2024ollivier} replaced the curvature term $\kappa_e(t)$ in the flow equation by the Lin-Lu-Yau Ricci curvature~\citep{lin2011ricci}, which is a limiting version of Ollivier-Ricci curvature. \cite{bai2024ollivier} established the local existence and uniqueness of solutions to this discrete Ricci flow. In particular, they proved the long-time existence (up to surgeries) of solutions under an exit condition.
More recently, \cite{ma2024modified} proposed a modified formulation in which the edge weight $w_e(t)$ on the right-hand side is replaced by the distance $\rho_e(t)$, namely,
\[
w_e^{\prime}(t) = -\kappa_e(t)\rho_e(t), \quad e \in \mathcal{E}.
\]
This formulation preserves the essential intuition of the Ricci flow. Along the flow, edges with positive curvature tend to shrink, while those with negative curvature tend to expand. At the same time, the initial value problem has a unique global solution without the exit condition. Therefore, in this work, we adopt this latter formulation as our discrete Ricci flow model for networks.

\section{Dependency Range of Ricci-Flow Geometry}
\label{app:dependency}
Let $\mathcal G$ be a graph in which two original edges are adjacent when they share an endpoint, and let
$B_r(e)$ be the radius-$r$ ball around edge $e$. Write one simultaneous flow update as
\[
\Phi_e(\mathbf w(t))=
w_e(t)-\delta\,\kappa_e(\mathbf w(t))\rho_e(\mathbf w(t)).
\]
For a target edge $e=(u,v)$, the measures $m_u(t)$ and $m_v(t)$ are normalized
using the current weights incident to $u$ and $v$. One update can therefore transmit dependence from edges in
$B_1(e)$. Composition of the simultaneous updates gives, by induction,
\[
\mathcal D^{w,\mathrm{loc}}_e(t)\subseteq B_t(e),
\qquad
\mathcal D^{\kappa,\mathrm{loc}}_e(t)\subseteq B_{t+1}(e),
\]
where $\mathcal D$ is the set of initial edge weights that may affect the
indicated quantity through this local channel.

Moreover, the complete Ollivier--Ricci calculation is not confined to this local
channel. Its transport cost matrix contains the shortest-path distance $d_t(a,b)$ for
$a\in\operatorname{supp}(m_u)$ and $b\in\operatorname{supp}(m_v)$, and
$\rho_e=d_t(u,v)$.  An edge on an active shortest path may therefore affect a
target curvature before the stepwise local channel reaches it.  This is why a
universal claim of exactly one additional hop per flow step would be too
strong.

These inclusions are possible-dependence bounds. Particular derivatives can
vanish because of symmetry, cancellation, or a change in the optimal transport basis.
We verified these two mechanisms on a 15-node weighted
path and a weighted 12-cycle, respectively.

The weighted path in Figure \ref{fig:dependency} gives an instance that attains
the successive boundaries. The initial weights
are $w_{(i,i+1)}=0.82+0.035i+0.025\sin(1.7i)$ for $i=0,\ldots,13$. The target edge
was $(7,8)$ and an initial edge weight at each line-graph distance
$r=1,\ldots,6$ was perturbed by $\epsilon=10^{-4}$. With $\alpha=0.5$ and
$\delta=0.12$, an effect was treated as nonzero when its finite-difference
magnitude $|\Delta q|/\epsilon$ exceeded $10^{-9}$. The first nonzero effects
were
\[
\begin{array}{c|rrrrrr}
	r & 1&2&3&4&5&6\\ \hline
	\text{target curvature }\kappa_e(t) &0&1&2&3&4&5\\
	\text{target weight }w_e(t)         &1&2&3&4&5&6
\end{array}
\]
as predicted by the local dependency channel. A perturbation at distance $r$ first
affects the target curvature at $t=r-1$ and the target weight at $t=r$. 

\begin{figure*}[htbp]
	\centering
	\begin{minipage}[t]{0.46\textwidth}
		\centering
		\resizebox{\linewidth}{!}{%
		\begin{tikzpicture}[x=0.42cm,y=0.42cm]
			\foreach \i in {0,...,14}{
				\node[circle,fill=black,inner sep=1.25pt,label=below:{\scriptsize\i}] (p\i) at (\i,0) {};
			}
			\foreach \i in {0,...,13}{
				\pgfmathtruncatemacro{\j}{\i+1}
				\draw[gray!65] (p\i)--(p\j);
			}
			\draw[red,very thick] (p7)--(p8);
			\draw[blue,very thick] (p13)--(p14);
			\node[red,above=3pt of p7.east,anchor=west] {\scriptsize target $(7,8)$};
			\node[blue,above=3pt of p13.east,anchor=west] {\scriptsize perturbed $(13,14)$};
		\end{tikzpicture}
		}
		
		\smallskip
		{\small (a) 15-node weighted path.}
	\end{minipage}
	\hfill
	\begin{minipage}[t]{0.46\textwidth}
		\centering
		\begin{tikzpicture}[scale=0.72]
			\foreach \i in {0,...,11}{
				\node[circle,fill=black,inner sep=1.25pt,label={[font=\scriptsize]90-30*\i:\i}]
				(c\i) at ({90-30*\i}:2.1cm) {};
			}
			\foreach \i in {0,...,10}{
				\pgfmathtruncatemacro{\j}{\i+1}
				\draw[gray!65] (c\i)--(c\j);
			}
			\draw[gray!65] (c11)--(c0);
			\draw[red,very thick] (c0)--(c1);
			\draw[blue,very thick] (c6)--(c7);
			\node[red] at (1.75,1.55) {\scriptsize target};
			\node[blue] at (-1.85,-1.55) {\scriptsize perturbation};
		\end{tikzpicture}
		
		{\small (b) Weighted 12-cycle.}
	\end{minipage}
	
	\caption{\textbf{Controlled geometric-dependency diagnostic. Red denotes the target edge and blue denotes the perturbed edge in (a) the 15-node weighted path and (b) the weighted 12-cycle.}}
	\label{fig:dependency}
\end{figure*}

For the weighted 12-cycle, the target was $(0,1)$ and the perturbed edge
$(6,7)$ was at line-graph distance six. The target has weight $1.5$, its two adjacent side edges have
weight $4.5$, and all other edges have weight $0.25$.
The perturbed edge lies on a shortest path
used in the target edge's optimal-transport cost matrix.  Its perturbation
changed the target curvature already at $t=0$ (finite-difference derivative
$-0.25$) and the target weight at $t=1$ (derivative $0.045$), confirming the
possible immediate non-local metric contribution.

\section{Mathematical Expressions of GEGCN}

\subsection{Geometric Evolution Generator via Discrete Ricci Flow}

Let $\mathcal{G}=(\mathcal{V},\mathcal{E})$ be an undirected graph with initial edge weight $w_e(0)$ for each edge $e\in\mathcal{E}$. The discrete Ricci flow is computed only on the original edge set $\mathcal{E}$. At time step $t$, the Ollivier-Ricci curvature $\kappa_e(t)$ and $\rho_{e}(t)$ of an edge $e=(u,v)$ are computed according to the current weights. The discrete Ricci flow evolves edge weights according to 
\begin{equation*}
	w_e(t+1)=w_e(t)-\delta\,\kappa_e(t)\rho_{e}(t), \quad t=0,\dots,T-1,
\end{equation*}
where $\delta>0$ is the step size. This process produces a curvature evolution sequence for each edge $e$,
\[
\mathcal{K}_e=[\kappa_e(0),\kappa_e(1),\dots,\kappa_e(T)],
\]
and a corresponding weight evolution sequence
\[
\mathcal{W}_e=[w_e(0),w_e(1),\dots,w_e(T)].
\]
After the Ricci-flow process is completed, we append one self-loop to each node to match the standard GCN design. Each appended self-loop is assigned the constant curvature sequence $[0,\dots,0]$ and the constant weight sequence $[1,\dots,1]$. These self-loops are used only when constructing $\tilde{\mathbf{A}}^*$ and do not affect the curvature, distance, or weight computation of the original edges.

\subsection{Structural Dynamics Encoding via LSTM}
\label{app:lstm}

For each edge $e\in\mathcal{E}$ (including the added self-loop on each node) and each time step $t$, the curvature and weight are concatenated to form the LSTM input $\mathbf{x}_e^{(t)}=\big[\kappa_e(t),\,w_e(t)\big]\in\mathbb{R}^2$.

The LSTM encoder updates the hidden state $\mathbf h_e^{(t)}$ and cell state $\mathbf c_e^{(t)}$ as
\begin{align*}
	\mathbf{f}_e^{(t)} &= \sigma\!\left(\mathbf{W}_f \mathbf{x}_e^{(t)} + \mathbf{U}_f \mathbf{h}_e^{(t-1)} + \mathbf{b}_f \right), \\
	\mathbf{i}_e^{(t)} &= \sigma\!\left(\mathbf{W}_i \mathbf{x}_e^{(t)} + \mathbf{U}_i \mathbf{h}_e^{(t-1)} + \mathbf{b}_i \right), \\
	\mathbf{o}_e^{(t)} &= \sigma\!\left(\mathbf{W}_o \mathbf{x}_e^{(t)} + \mathbf{U}_o \mathbf{h}_e^{(t-1)} + \mathbf{b}_o \right), \\
	\tilde{\mathbf{c}}_e^{(t)} &= \tanh\!\left(\mathbf{W}_c \mathbf{x}_e^{(t)} + \mathbf{U}_c \mathbf{h}_e^{(t-1)} + \mathbf{b}_c \right), \\
	\mathbf{c}_e^{(t)} &= \mathbf{f}_e^{(t)}\odot \mathbf{c}_e^{(t-1)} + \mathbf{i}_e^{(t)}\odot \tilde{\mathbf{c}}_e^{(t)}, \\
	\mathbf{h}_e^{(t)} &= \mathbf{o}_e^{(t)}\odot \tanh\!\big(\mathbf{c}_e^{(t)}\big),
\end{align*}
where $\sigma(\cdot)$ denotes the sigmoid function and $\odot$ is the element-wise product.
The cell state $\mathbf c_e^{(t)}$ is the gated internal memory that
combines the retained previous state with the current candidate state. It can
therefore carry selected geometric information across Ricci-flow steps.
We input the final hidden state $\mathbf{h}_e^{(T)}$ of each edge into an edge-scoring head as follows:
\begin{equation*}
	a^*_e = \sigma\left( \mathbf{W}_s \mathbf{h}_e^{(T)} + b_s \right),
\end{equation*}
where $a^*_e \in (0, 1)$ is taken as the edge importance score of edge $e$.

\subsection{Geometry-aware Adjacency Normalization}

Let $\tilde{\mathbf{A}}^*\in\mathbb{R}^{N\times N}$ denote the edge importance matrix,
\[
\tilde{\mathbf{A}}^*_{ij} =
\begin{cases}
	a_{e}^*, & e=(v_i,v_j)\in\mathcal{E},\ \text{or}\ e=(v_i,v_i) \ \text{is a self-loop},\\
	0, & \text{otherwise}.
\end{cases}
\]
The corresponding degree matrix is
\[
\tilde{\mathbf{D}}^*=\mathrm{diag}\!\left(\sum_j \tilde{\mathbf{A}}^*_{ij}\right).
\]
The geometry-aware normalized adjacency matrix is then defined as
\begin{equation*}
	\hat{\mathbf{A}}^*=\tilde{\mathbf{D}}^{*-\frac12}\tilde{\mathbf{A}}^*\tilde{\mathbf{D}}^{*-\frac12}.
\end{equation*}

\subsection{GEGCN Forward Propagation}

Given input node features $\mathbf{H}^{(0)}=\mathbf{X}$, the $\ell$-th layer of GEGCN updates node representations according to
\begin{equation*}
	\mathbf{H}^{(\ell+1)}=\sigma\!\left(\hat{\mathbf{A}}^*\mathbf{H}^{(\ell)}\mathbf{W}^{(\ell)}\right),
	\quad \ell=0,\dots,L-1.
\end{equation*}
The final node embeddings $\mathbf{H}^{(L)}$ are used for downstream learning tasks.

\section{Details of Experiments}

\subsection{Environment}
To ensure the reproducibility of our results, we provide the details of the experimental environment. All simulations and model training were executed on a Google Colab Pro+ environment with the following specifications:

\begin{itemize}
	\item \textbf{GPU:} A single NVIDIA A100-SXM4 GPU with 40GB of high-bandwidth memory (HBM2). This hardware supports the TensorFloat-32 (TF32) precision format. 
	\item \textbf{CPU:} An Intel(R) Xeon(R) CPU running at 2.20GHz (Cascade Lake architecture). The system provides 12 logical cores, supporting the AVX-512 instruction set.
	\item \textbf{Software Stack:} The models were implemented using Python 3.10.12 and PyTorch 2.x. Graph-related operations and message-passing layers were built upon PyTorch Geometric (PyG) 2.4.0. CUDA 12.4 was used as the backend for GPU acceleration.
\end{itemize}

\subsection{Hyperparameter Settings}\label{app:hyper_setting}
To ensure the robustness and strong performance of the proposed GEGCN, we conducted automated hyperparameter optimization using Optuna. For each dataset, we performed up to 100 trials over the search space summarized in Table~\ref{tab:hyperparams}. The final configuration for each benchmark was selected based on the highest mean validation accuracy.

\begin{table}[H]
	\centering
	\caption{Hyperparameter Search Space and Optimization Details.}
	\label{tab:hyperparams}
	\footnotesize
	\renewcommand{\arraystretch}{1.2}
	\begin{tabularx}{\columnwidth}{@{}lXl@{}}
		\toprule
		\textbf{Hyperparameter} & \textbf{Search Range/Values} & \textbf{Distribution} \\ 
		\midrule
		\rowcolor{gray!10} \multicolumn{3}{l}{\textit{Optimization Parameters}} \\
		Learning rate ($\eta$) & $[10^{-5}, 10^{-2}]$ & Log-uniform \\
		Weight decay ($\lambda$) & $[10^{-6}, 10^{-2}]$ & Log-uniform \\
		Dropout rate ($p$) & $[0.01, 0.99]$ & Uniform \\
		Hidden dimension ($d$) & $\{32,64, 128\}$ & Categorical \\
		\midrule
		\rowcolor{gray!10} \multicolumn{3}{l}{\textit{Curvature \& Flow Parameters}} \\
		Flow steps ($T$) & $\{1, 2, \dots, 10\}$ & Discrete Uniform \\ 
		Step size ($\delta$) & $[0.01, 1.0)$ & Uniform \\
		Random walk parameter ($\alpha$) & $ \{0,0.1,\cdots,1\}$ & Discrete Uniform \\
		Sinkhorn regularization ($\epsilon$) & 0.1 & Fixed \\
		\bottomrule
	\end{tabularx}
\end{table}

\subsection{Dataset Statistics}
Let $\mathcal{G}=(\mathcal{V},\mathcal{E})$ be a graph with nodes $\mathcal{V}$ and edges $\mathcal{E}$. Each node $v$ has a class label $y_v \in \{1, \dots, m\}$. We denote by $d(v)$ the degree of $v$, that is $d(v) = |\mathcal{N}(v)|$. The homophily index $\mathcal{H}_{\mathcal{G}} \in [0, 1]$, as proposed by \cite{peigeomgcn2020}, is defined as:
\begin{equation*}
	\mathcal{H}_{\mathcal{G}} = \dfrac{1}{|\mathcal{V}|}\sum_{v\in \mathcal{V}}\dfrac{|\{u\in \mathcal{N}(v) : y_u = y_v\}|}{d(v)}.
\end{equation*}
The detailed homophily index and statistics of the datasets are summarized in Table \ref{tab:data}.

\begin{table*}[!t]
	\centering
	\caption{Statistics of Datasets.}
	\label{tab:data}
	\footnotesize
	\begin{tabular}{lccccc}
		\toprule
		\textbf{Dataset} & \textbf{Nodes} & \textbf{Edges} & \textbf{Features} & \textbf{Classes} & \textbf{Homophily} ($\mathcal{H}(G)$) \\
		\midrule
		Cora        & 2,708  & 5,278   & 1,433 & 7  & 0.81 \\
		Citeseer    & 3,327  & 4,552   & 3,703 & 6  & 0.74 \\
		PubMed      & 19,717 & 44,324  & 500   & 3  & 0.80 \\
		Amazon Photos       & 7,650  & 119,081 & 745   & 8  & 0.83 \\
		Coauthor CS & 18,333 & 81,894  & 6,805 & 15 & 0.81 \\
		Cornell   & 183   & 277    & 1,703 & 5 & 0.12 \\
		Texas     & 183   & 279    & 1,703 & 5 & 0.06 \\
		Wisconsin & 251   & 450    & 1,703 & 5 & 0.17 \\
		Chameleon & 2,277 & 31,371 & 2,325 & 5 & 0.23 \\
		Actor     & 7,600 & 26,659 & 932   & 5 & 0.22 \\
		Chameleon-f & 890 & 8,854 & 2,325 & 5 & 0.24 \\
		Squirrel-f & 2,223 & 46,998 & 2,089 & 5 & 0.21 \\
		Roman-empire & 22,662 & 32,927 & 300 & 18 & 0.05 \\
		Amazon-ratings & 24,492 & 93,050 & 300 & 5 & 0.38 \\
		Minesweeper & 10,000 & 39,402  & 7     & 2  & 0.68 \\
		Questions   & 48,921 & 153,540 & 301   & 2  & 0.84 \\
		OGBN-arXiv  & 169,343 & 1,166,243 & 128   & 40  & 0.64 \\
		\bottomrule
	\end{tabular}
\end{table*}

\subsection{Precomputation Details}\label{time}

As shown in Table~\ref{tab:precompute_time}, the discrete Ricci flow is precomputed offline as a one-time preprocessing step. 
Although it involves iterative Sinkhorn updates, it introduces no additional latency during model training or inference. 
The preprocessing cost remains moderate across the benchmark datasets. 
Even for relatively dense graphs such as Amazon Photos, computing $T=10$ Ricci-flow steps takes about 30 minutes. For the large-scale OGBN-arXiv benchmark, with approximately 9.8 times more edges than Amazon Photos, the preprocessing time rises to roughly 7.5 times that of Amazon Photos, yet this overhead remains confined to the offline stage and does not enter the training phase. This cost is paid only once and can be reused across different training runs, hyperparameter settings, and random data splits. 
In practice, this phase can be further accelerated through engineering optimizations, since curvature updates across edges are naturally independent and therefore well suited for parallelization on multi-core CPUs or GPU clusters.

\begin{table*}[!t]
	\centering
	\caption{Efficiency of offline precomputation. We report the estimated execution time (in seconds) for computing $T=10$ steps of discrete Ricci flow.}
	\label{tab:precompute_time}
	\footnotesize
	\setlength{\tabcolsep}{5pt}
	\renewcommand{\arraystretch}{1.0}
	\begin{minipage}[t]{0.47\textwidth}
		\centering
		\begin{tabular}{lrrr}
			\toprule
			\textbf{Dataset} & \textbf{Nodes} & \textbf{Edges} & \textbf{Time (s)} \\
			\midrule
			Cora            & 2,708  & 5,278   & 34.90  \\
			Citeseer        & 3,327  & 4,552   & 2.59   \\
			PubMed          & 19,717 & 44,324  & 851.70 \\
			Amazon Photos   & 7,650  & 119,081 & 1831.04 \\
			Coauthor CS     & 18,333 & 81,894  & 1079.04 \\
			Cornell         & 183    & 277     & 4.84   \\
			Texas           & 183    & 279     & 3.17   \\
			Wisconsin       & 251    & 450     & 5.15   \\
			\bottomrule
		\end{tabular}
	\end{minipage}\hfill
	\begin{minipage}[t]{0.47\textwidth}
		\centering
		\begin{tabular}{lrrr}
			\toprule
			\textbf{Dataset} & \textbf{Nodes} & \textbf{Edges} & \textbf{Time (s)} \\
			\midrule
			Chameleon       & 2,277  & 31,371  & 745.26 \\
			Actor           & 7,600  & 26,659  & 173.01 \\
			Chameleon-f     & 890    & 8,854   & 110.74 \\
			Squirrel-f      & 2,223  & 46,998  & 2654.67 \\
			Roman-empire    & 22,662 & 32,927  & 8.78   \\
			Amazon-ratings  & 24,492 & 93,050  & 169.73 \\
			Minesweeper     & 10,000 & 39,402  & 77.30  \\
			Questions       & 48,921 & 153,540 & 191.13 \\
			OGBN-arXiv 	    & 169,343 & 1,166,243 & 13,763.28 \\
			\bottomrule
		\end{tabular}
	\end{minipage}
	\vspace{-0.8em}
\end{table*}

\subsection{Reproducibility Protocol}\label{app:reproducibility}

\paragraph{Preprocessing and data alignment.}
The geometric evolution generator uses graph topology and edge weights, without class labels or node features. For a fixed graph and preprocessing configuration, it computes the ordered pairs $[\kappa_e(t),w_e(t)]$ for $t=0,\ldots,T$ on the edge set. Edge identifiers remain consistent across all flow states and when the encoded scores are assembled into the propagation matrix. As specified in Sections~\ref{generator} and~\ref{lstm}, one self-loop per node is appended only after Ricci-flow preprocessing, with constant curvature $0$ and weight $1$ throughout the sequence. These self-loops do not participate in the curvature or shortest-path computations.

\paragraph{Evaluation protocol.}
Section~\ref{exp_data} specifies the dataset splits: 20\%/10\%/70\% for the five homophilic datasets, 60\%/20\%/20\% with a fixed test set for the five traditional heterophilic datasets, the cited benchmark splits for the six filtered heterophilic datasets, and the official temporal split for OGBN-arXiv. Hyperparameters are selected using mean validation accuracy over the search spaces in Table~\ref{tab:hyperparams}, with up to 100 Optuna trials per dataset. Test labels are reserved for final evaluation. Appendix D.1 records the experimental environment. The source-code repository is provided in the Data and code availability statement.


\end{document}